\documentclass[11pt,a4paper]{article}
\usepackage[hyperref]{acl2018}
\usepackage{times}
\usepackage{latexsym}

\usepackage{url}
\usepackage{graphicx}  %Required

\usepackage{amsmath,amsfonts,amsthm,amssymb}
\usepackage{mathtools}
\newtheorem{definition}{Definition}
\newtheorem{theorem}{Theorem}
\newtheorem{corollary}{Corollary}[theorem]

\newtheorem{proposition}{Proposition}
\usepackage{subfigure}
\usepackage{algpseudocode}
\usepackage{algorithm}
\usepackage{enumitem}
\usepackage{pifont}
\usepackage{booktabs}

\DeclareMathOperator*{\argmax}{argmax}

\newcommand*{\QED}{\hfill\ensuremath{\square}}
\newcommand{\fw}{0.23}
\newcommand{\cmark}{\ding{51}}%
\newcommand{\xmark}{\ding{55}}%

\aclfinalcopy % Uncomment this line for the final submission
\title{Active Learning of Strict Partial Orders: A Case Study on Concept Prerequisite Relations}

\author{
Chen Liang$^{1}$ $\quad$ Jianbo Ye$^{1}$ $\quad$ Han Zhao$^{2}$ $\quad$ Bart Pursel$^{1}$ $\quad$ C. Lee Giles$^{1}$\\
	$^{1}$The Pennsylvania State University\\
   $^{2}$Carnegie Mellon University\\
  {\tt \{cul226, jxy198, bkp10, giles\}@ist.psu.edu}\\
  {\tt han.zhao@cs.cmu.edu}\\
}

\date{}

\begin{document}
\maketitle
\begin{abstract}
Strict partial order is a mathematical structure commonly seen in relational data.
One obstacle to extracting such type of relations at scale is the lack of large scale labels
for building effective data-driven solutions. 
We develop an active learning framework for mining such relations subject to a strict order.
Our approach incorporates relational reasoning not only in finding new unlabeled pairs
whose labels can be deduced from an existing label set, but also in devising
new query strategies that consider the relational structure of labels.
Our experiments on concept prerequisite relations show our proposed framework can substantially improve
the classification performance with the same query budget compared to other baseline approaches.
\end{abstract}

\section{Introduction}
Pool-based active learning is a learning framework where the learning algorithm is allowed to access a set of unlabeled examples and ask for the labels of any of these examples~\cite{angluin1988queries,cohn1996active,settles2010active}.
Its goal is to learn a good classifier with significantly fewer labels by actively directing the queries to the most ``valuable'' examples.
In a typical setup of active learning, the label dependency among labeled or unlabeled examples is not considered.
But data and knowledge in the real world are often embodied with prior relational structures.
Taking into consideration those structures in building machine learning solutions can be necessary and crucial~\cite{getoor2007introduction}.
The goal of this paper is to investigate the query strategies in active learning of a strict partial order, 
namely, when the ground-truth labels of examples constitute an irreflexive and transitive relation.
In this paper, we develop efficient and effective algorithms extending popular query strategies used in active learning to work with such relational data.
%Strict partial order provides a mathematical representation for building domain specific knowledge graphs.
%The creation of such graphs facilitates the processes of automatic relational inference and knowledge management, without which domain experts are instead needed to manually handle this work.
%Active learning provides an opportunity to semi-automatically create such knowledge graphs by incorporating standard statistical learning in a cheaper way, therefore, should be of interest to practitioners from different domains.
We study the following problem in the active learning context:

\noindent\textbf{Problem.} Given a finite set $V$, a strict order on $V$ is a type of irreflexive and transitive (pairwise) relation. Such a strict order is represented by a subset $G \subseteq V\times V$. {Given an unknown strict order $G$, an oracle $W$ that returns $W(u,v) = -1 + 2 \cdot \mathbf 1 [(u,v) \in G] \in \{-1,1\}$, and a feature extractor $\mathcal F: V\times V \mapsto \mathbb R^d$, find $h: \mathbb R^d \mapsto \{-1,1\}$ from a hypothesis class $\mathcal H$ that predicts whether or not $(u,v)\in G$ for each pair $(u,v) \in V\times V$ and $u\neq v$ (using $h(\mathcal F(u,v))$) by querying $W$ a finite number of $(u,v)$ pairs from $V\times V$. }

Our main focus is to develop reasonable query strategies in active learning of a strict order exploiting both the knowledge from (non-consistent) classifiers trained on a limited number of labeled examples and the deductive structures among pairwise relations.
Our work also has a particular focus on {\em partial orders}. If the strict order is total, a large school called ``learning to rank'' has studied this topic~\cite{burges2005learning,liu2009learning}, some of which are under the active learning setting~\cite{donmez2008optimizing,long2010active}.
Learning to rank relies on binary classifiers or probabilistic models that are consistent with the rule of a total order. Such approaches are however limited in a sense to principally modeling a partial order: a classifier consistent with a total order will always have a non-zero lower bound of error rate, if the ground-truth is a partial order but not a total order.

In our active learning problem,
incorporating the deductive relations of a strict order in soliciting examples to be labeled is non-trivial and important.
The challenges motivating us to pursue this direction can be explained in three folds:
First, any example whose label can be deterministically reasoned from a labeled set by using the properties of strict orders does not need further manual labeling or statistical prediction.
Second, probabilistic inference of labels based on the independence hypothesis, as is done in the conventional classifier training, is {\em not} proper any more because the deductive relations make the labels of examples dependent on each other.
Third, in order to quantify how valuable an example is for querying, one has to combine uncertainty and logic to build proper representations. Sound and efficient heuristics with empirical success are to be explored.

One related active learning work that deals with a similar setting to ours is~\cite{rendle2008active}, whereas equivalence relations are considered instead. Particularly, they made several crude approximations in order to expedite the expected error calculation to a computational tractable level. We approach the design of query strategies from a different perspective while keeping efficiency as one of our central concerns.

To empirically study the proposed active learning algorithm, we apply it to \textit{concept prerequisite learning problem}~\cite{talukdar2012crowdsourced,DBLP:conf/emnlp/LiangWHG15}, where the goal is to predict whether a concept $A$ is a prerequisite of a concept $B$ given the pair $(A,B)$. Although there have been some research efforts towards learning prerequisites~\cite{vuong2010method,talukdar2012crowdsourced,DBLP:conf/emnlp/LiangWHG15,wang2016using,scheines2014discovering,liu2016learning, pan2017prerequisite}, the mathematical nature of the prerequisite relation as strict partial orders has not been investigated. In addition, one obstacle for effective learning-based solutions to this problem is the lack of large scale prerequisite labels.
\citet{liang2018investigating} applied standard active learning to this problem without utilizing relation properties of prerequisites.
Active learning methods tailored for strict partial orders provide a good opportunity to tackle the current challenges of concept prerequisite learning.

Our main contributions are summarized as follows:
First, we propose a new efficient reasoning module for monotonically calculating the deductive closure  under the assumption of a strict order. This computational module can be useful for general AI solutions that need fast reasoning in regard to strict orders.
Second, we apply our reasoning module to extend two popular active learning approaches to handle relational data and empirically achieve substantial improvements. This is the first attempt to design active learning query strategies tailored for strict partial orders.
Third, under the proposed framework, we solve the problem of concept prerequisite learning and our approach appears to be successful on data from four educational domains, whereas previous work have not exploited the relational structure of prerequisites as strict partial orders in a principled way.

\section{Reasoning of a Strict Order}
\label{sec:spo}

\begin{figure*}[ht!]
\centering
% \subfigure[]{
\includegraphics[width=\fw\textwidth,trim=5cm 5cm 6cm 5cm]{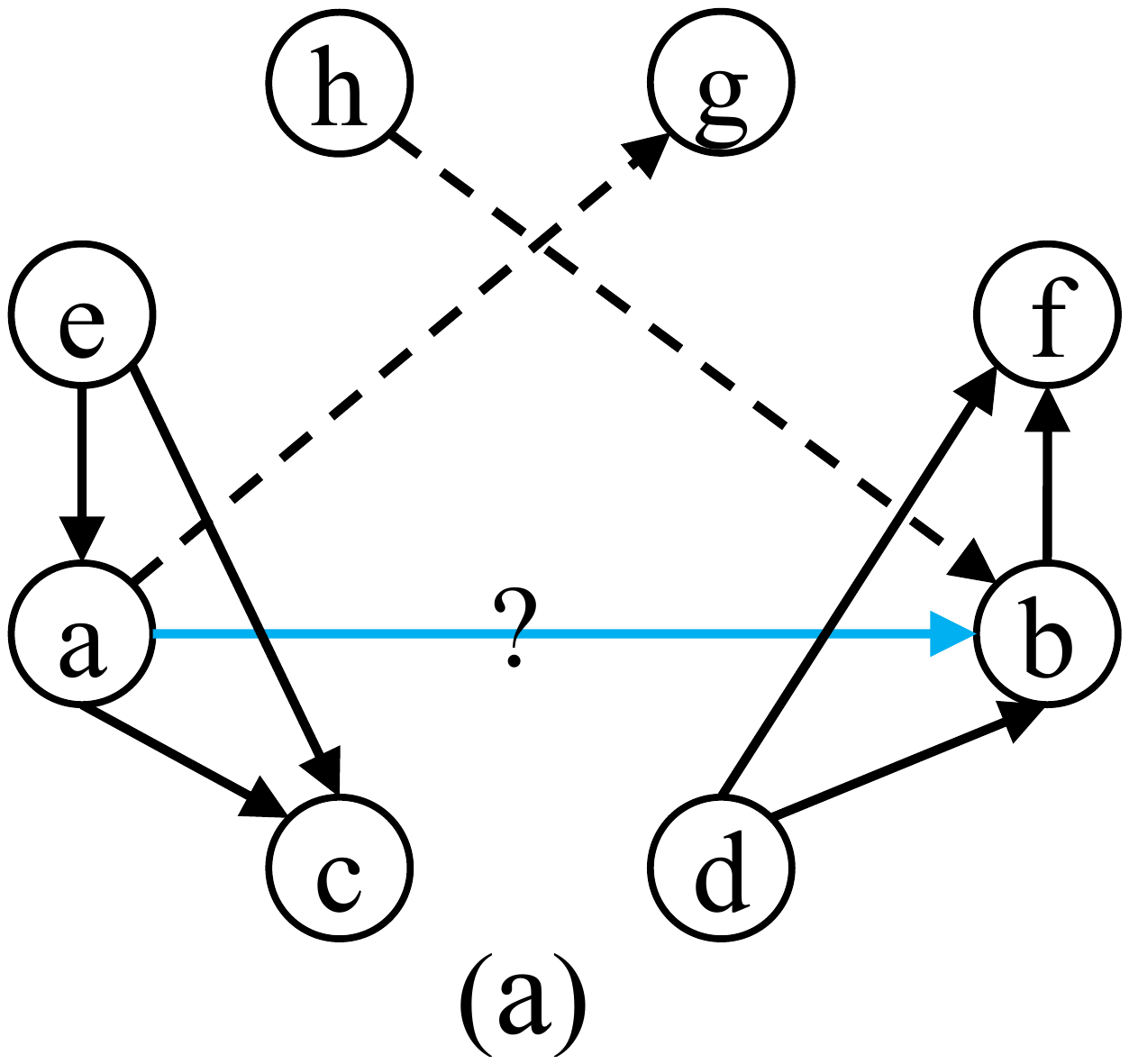}
% }
% \subfigure[]{
\includegraphics[width=\fw\textwidth,trim=5cm 5cm 6cm 5cm]{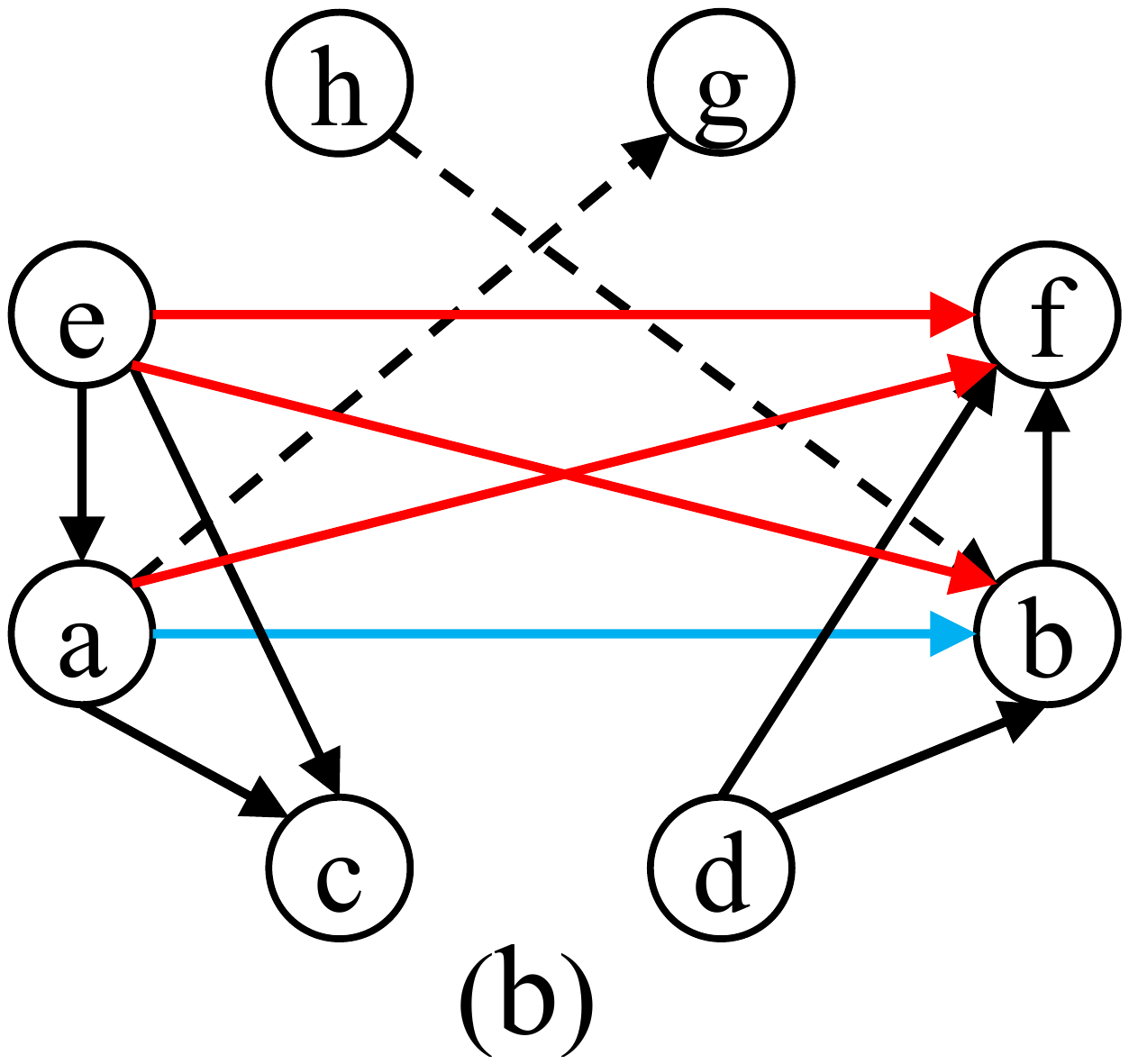}
% }
% \subfigure[]{
\includegraphics[width=\fw\textwidth,trim=5cm 5cm 6cm 5cm]{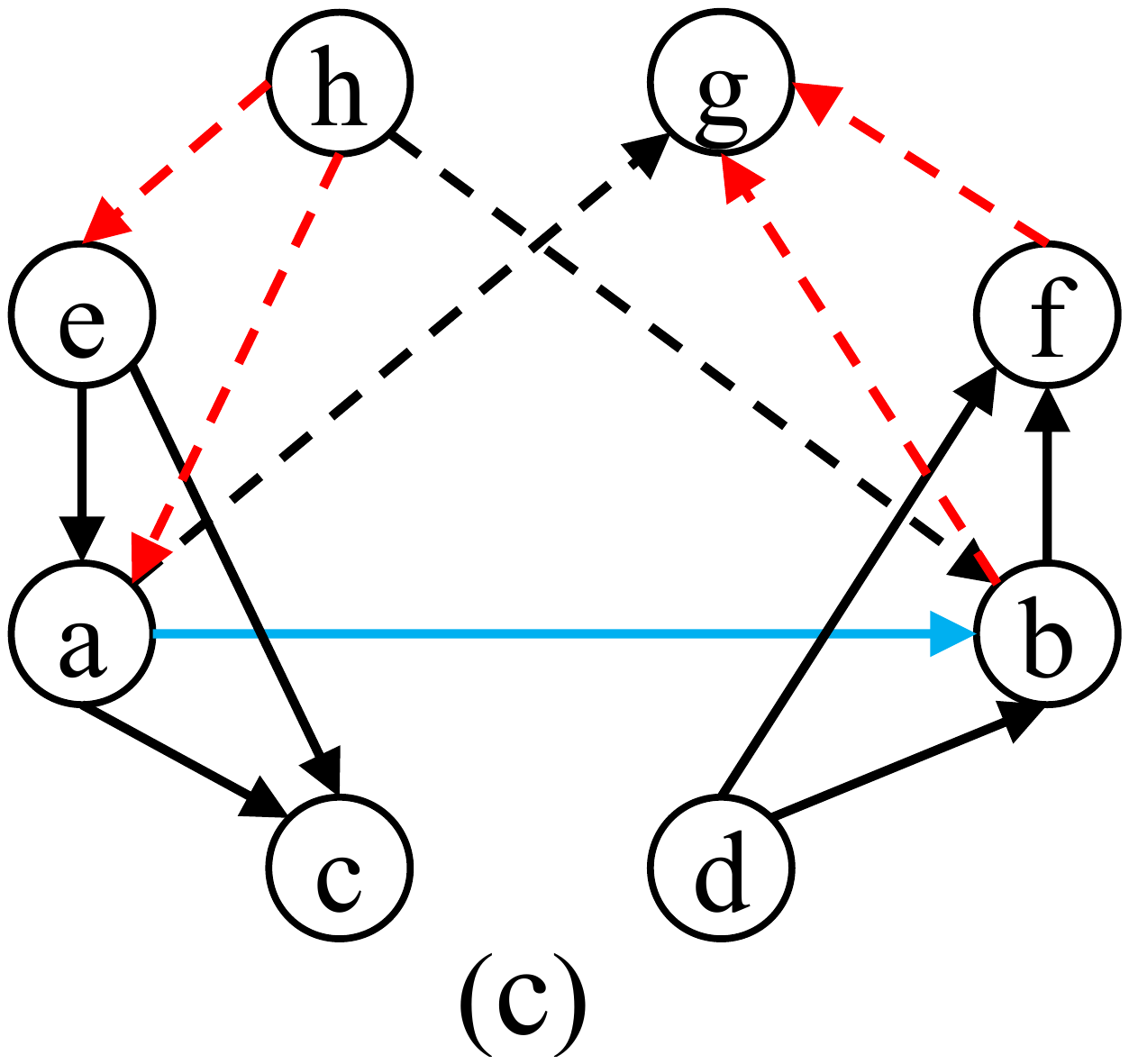}
% }
% \subfigure[]{
\includegraphics[width=\fw\textwidth,trim=5cm 5cm 6cm 5cm]{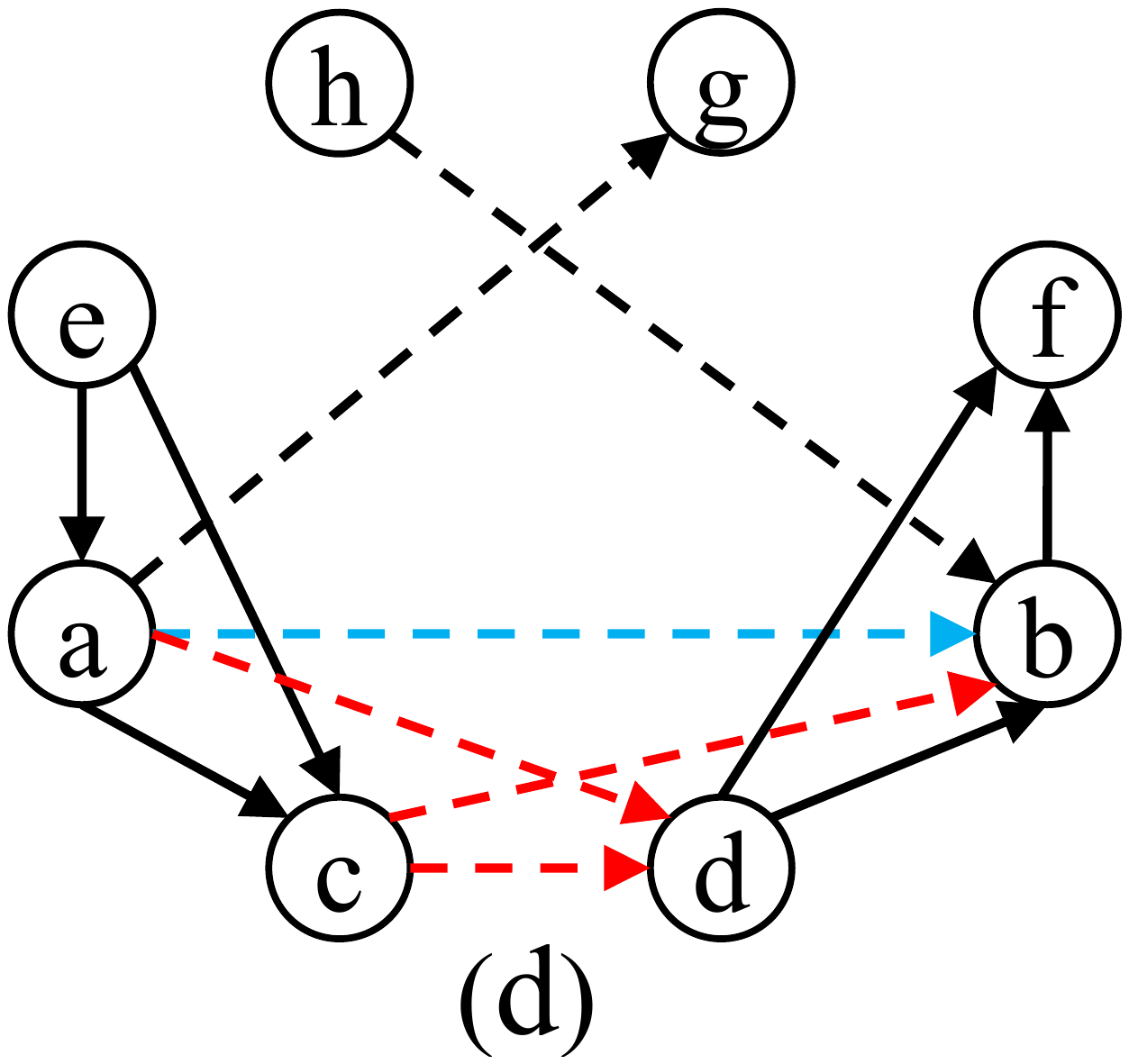}
% }

\caption{\small Following the notations in Theorem~\ref{thm:cxy}:
(a) Black lines are pairs in $H$, solid lines are pairs in $G$, and dashed lines are pairs in $G^c$.
The pair $(a,b)$ in the cyan color is the pair to be labeled or deduced.
(b) If $(a,b)\in G$, $\{(a,b),(e,f),(a,f),(e,b)\} \subseteq N_{(a,b)}$.
(c) If $(a,b)\in G$, $\{(h,e),(h,a)\}\subseteq T_{(a,b)}$ and $\{(b,g),(f,g)\}\subseteq S_{(a,b)}$.
(d) If $(a,b)\in G^c$, $\{(a,b),(a,d),(c,b),(c,d)\}\subseteq N'_{(a,b)}$. Likewise, if $\exists (x,y)\in G, \mbox{s.t.} (a,b)\in S_{(x,y)}\cup T_{(x,y)}$, $\{(a,b),(a,d),(c,b)\}\subseteq O_{(x,y)}$.}
\label{fig:reasoning}
\end{figure*}

\subsection{Preliminary}
\begin{definition}[Strict Order]
\label{def:sorder}
Given a finite set $V$, a subset $G$ of $V\times V$ is called a strict order if and only if it satisfies the two conditions:
(i) if $(a,b)\in G$ and $(b,c)\in G$, then $(a,c)\in G$; (ii) if $(a,b)\in G$, then $(b,a)\not\in G$.
\end{definition}

\begin{definition}[$G$-Oracle]
For two subsets $G, H\subseteq V\times V$, a function denoted as
$W_H(\cdot,\cdot):H \mapsto \{-1,1\}$ is called a $G$-oracle on $H$ iff
for any $(u,v)\in H$, $W_H(u,v)=-1 + 2 \cdot \mathbf 1[(u,v)\in G] $.
\end{definition}
% A strict order $G$ is a binary relation that is both transitive and irreflexive. 
The $G$-oracle returns a label denoting whether a pair belongs to $G$.

% Moreover, if $G$ is a strict order of $V$, for any $H\subseteq V\times V$, the $G$-oracle $W_H$
% is called {\em an oracle of strict order}.

\begin{definition}[Completeness of an Oracle]\label{def:complete}
A $G$-oracle of strict order $W_H$ is called complete if and only if $H$ satisfies:
for any $a,b,c\in V$,
(i) if $(a,b)\in H\cap G$, $(b,c)\in H\cap G$, then $(a,c)\in H\cap G$;
(ii) if $(a,b)\in H\cap G$, $(a,c)\in H\cap G^c$, then $(b,c)\in H\cap G^c$;
(iii) if $(b,c)\in H\cap G$, $(a,c) \in H\cap G^c$, then $(a,b)\in H\cap G^c$;
(iv) if $(a,b)\in H\cap G$, then $(b,a)\in H\cap G^c$,
where $G^c$ is the complement of $G$.
\end{definition}
$W_H$ is called complete if it is consistent under transitivity when restricted on pairs from $H$.

\begin{definition}[Closure]~\label{def:closure}
Given a strict order $G$, for any $H\subseteq V\times V$, its closure is defined to be the
smallest set $\overline H$ such that $H\subseteq \overline H$ and the G-oracle $W_{\overline H}$ is complete.
\end{definition}

\begin{proposition}~\label{prop:uniq}
For any $H\subseteq V\times V$,
the closure of $H$ subject to a strict order $G$ is unique.
\footnote{Please see the supplemental material for the proofs of propositions and theorems introduced hereafter.}
\end{proposition}

\begin{proposition}~\label{prop:first}
Let $G$ be a strict order of $V$.
For a complete $G$-oracle $W_H$, $H\cap G$ is also a strict order of $V$.
% (See supplemental material for the proof.)
\end{proposition}

\begin{definition}[Descendant and Ancestor]
Given a strict order $G$ of $V$ and $a\in V$, its ancestor subject to $G$
is $A_a^G:=\{b \mid (b,a)\in G\}$ and its descendant is $D_a^G:=\{b \mid (a,b)\in G\}$.
\end{definition}

\subsection{Reasoning Module for Closure Calculation}
With the definitions in the previous section, this section proposes a reasoning module that is designed to monotonically calculate the deductive closure for strict orders.
Remark that a key difference between the traditional transitive closure and our definition of closure (Definition~\ref{def:complete}\&\ref{def:closure}) is that the former only focuses on $G$ but the latter requires calculation for both $G$ and $G^c$.
In the context of machine learning, relations in $G$ and $G^c$ correspond to positive examples and negative examples, respectively.
Since both of these examples are crucial for training classifiers, existing algorithms for calculating transitive closure such as the Warshall algorithm are not applicable.
Thus we propose the following theorem for monotonically computing the closure.

\begin{theorem}
\label{thm:cxy}
Let $G$ be a strict order of $V$ and $W_H$ a complete $G$-oracle on $H\subseteq V\times V$.
For any pair $(a,b)\in V\times V$, define the notation $C_{(a,b)}$ by
\small
\begin{enumerate}
\item[(i)] If $(a,b)\in H$, $C_{(a,b)} := H$.
\item[(ii)] If $(a,b)\in G^c\cap H^c$, $C_{(a,b)} := H \cup N'_{(a,b)}$ where
\[N'_{(a,b)} := \{(d,c)| c\in A_b^{G\cap H} \cup \{b\}, d\in D_a^{G\cap H} \cup \{a\}\},\]
and particularly $N'_{(a,b)}\subseteq G^c$.
\item[(iii)] If $(a,b)\in G\cap H^c$, $C_{(a,b)} := H \cup N_{(a,b)} \cup R_{(a,b)}\cup S_{(a,b)} \cup T_{(a,b)}\cup O_{(a,b)} $, where
{
\allowdisplaybreaks
\begin{eqnarray*}
\!\!\!\!\!\!N_{(a,b)} &\!\!\!\!:=\!\!\!\!& \{(c,d)\mid c\in A_a^{G\cap H}\cup \{a\}, d\in D_b^{G\cap H} \cup \{b\}\},\\
\!\!\!\!\!\!R_{(a,b)} &\!\!\!\!:=\!\!\!\!& \{(d,c) \mid (c,d) \in N_{(a,b)}\},\\
\!\!\!\!\!\!S_{(a,b)} &\!\!\!\!:=\!\!\!\!& \{(d,e) \mid c\in A_{a}^{G\cap H}\cup \{a\}, d \in D_{b}^{G\cap H}\cup \{b\}, \\
&&  ~~ (c,e) \in {G^c\cap H}\}, \\
\!\!\!\!\!\!T_{(a,b)} &\!\!\!\!:=\!\!\!\!& \{(e,c) \mid c\in A_{a}^{G\cap H}\cup \{a\}, d \in D_{b}^{G\cap H}\cup \{b\}, \\
&&  ~~ (e,d) \in {G^c\cap H}\}, \\
\!\!\!\!\!\!O_{(a,b)} &\!\!\!\!:=\!\!\!\!& \bigcup\nolimits_{(c,d)\in S_{(a,b)}\cup T_{(a,b)}} N''_{(c,d)},\\
\!\!\!\!\!\!N''_{(c,d)} &\!\!\!\!:=\!\!\!\!& \{(f,e) \mid e\in A_d^{G\cap (H\cup N_{(a,b)})} \cup \{d\}, \\
&& ~~ f\in D_c^{G\cap (H\cup N_{(a,b)})} \cup \{c\}\}.
\end{eqnarray*}
}
In particular, $N_{(a,b)}\subseteq G$ and $R_{(a,b)}\cup S_{(a,b)} \cup T_{(a,b)}\cup O_{(a,b)}\subseteq G^c$.
\end{enumerate}
For any pair $(x,y)\in V\times V$, the closure of $H'=H\cup \{(x,y)\}$ is $C_{(x,y)}$.
% (Refer to Appendix~\ref{sec:proof} for the proof. Also, see Fig.~\ref{fig:reasoning} for the explanation of each equation.)
\end{theorem}

% Please refer to the supplemental material for the proof. 
% In addition, 
Figure~\ref{fig:reasoning} provides an informal explanation of each necessary condition (except for $R_{(a,b)}$) mentioned in the theorem.
%Note that $C_{(a,b)}$ can still be calculated given $H$ without knowing all elements in $G$ as long as one assumes whether or not $(a,b)\in G$.
%Such observation is important for developing our active learning framework and its query strategies.
If $(a,b)$ is a positive example, i.e. $(a,b)\in G$, then (i) $N_{(a,b)}$ is a set of inferred positive examples by transitivity; (ii) $R_{(a,b)}$ is a set of inferred negative examples by irreflexivity; (iii) $S_{(a,b)}$ and $T_{(a,b)}$ are sets of inferred negative examples by transitivity; (iv) $O_{(a,b)}$ is a set of negative examples inferred from $S_{(a,b)}$ and $T_{(a,b)}$.
If $(a,b)$ is a negative example, i.e. $(a,b)\in G^c$, then $N'_{(a,b)}$ is a set of negative examples inferred by transitivity.

\subsection{Computational Efficiency}
As we will elaborate later, one computational hurdle of our active learning algorithm
is to efficiently calculate the closure set
$C_{(x,y)}$ given a complete $G$-oracle $W_H$. In particular, among all the formula in Theorem~\ref{thm:cxy}, we found the main bottleneck is to efficiently calculate $O_{(x,y)}$ whose worst time complexity is $O(\left|H\right|^3)$ (followed by Prop.~\ref{prop:complexity}), while others can be done in $O(\left|H\right|^2)$.
\begin{proposition}
\label{prop:complexity}
Following the notations in Theorem~\ref{thm:cxy}, 
given $S_{(a,b)}\cup T_{(a,b)}$, the worst time complexity 
to calculate $O_{(a,b)}$ is $O(|H|^3)$. 
% (See supplemental material for the proof.)
\end{proposition}

Using Prop.~\ref{prop:efficiency},
one can show that there exists a pruning rule that cuts a major proportion of
redundant set operations in calculating $O_{(a,b)}$.
\begin{proposition}
\label{prop:efficiency}
Following the notations in Theorem~\ref{thm:cxy}, we have $N''_{(c',d')} \subseteq N''_{(c,d)}$ if $(c',d')\in N''_{(c,d)}$. 
% (See supplemental material for the proof.)
\end{proposition}
We also conduct empirical studies to examine the growth rate of 
calculating $C_{(x,y)}$. In practice, we find the empirical growth rate is closer 
to linear rate, which means the worst time complexity bound presented here is very conservative.

\section{Pool-Based Active Learning}
\label{sec:al}
The pool-based sampling~\cite{lewis1994sequential} is a typical active learning scenario in which one maintains a labeled set $\mathcal D_l$ and an unlabeled set $\mathcal D_u$. In particular, we let
$\mathcal D_u \cup \mathcal D_l = \mathcal D=\{1,\ldots, n\}$ and  $\mathcal D_u \cap \mathcal D_l = \varnothing$.
For $i\in\{1,\ldots, n\}$, we use $\mathbf{x}_i\in \mathbb R^d$ to denote a feature vector representing the $i$-th instance, and $y_i\in\{-1,+1\}$ to denote its groundtruth class label.
At each round, one or more instances are selected from $\mathcal D_u$ whose label(s) are then requested, and the labeled instance(s) are then moved to $\mathcal D_l$. Typically instances are queried in a prioritized way such that one can obtain good classifiers trained with a substantially smaller set $D_l$. We focus on the pool-based sampling setting where queries are selected in serial, i.e., one at a time. 
% Algorithm~\ref{algo:al} presents the typical setting of serial pool-based active learning.

% \begin{algorithm}[t]
% \caption{Pseudocode for pool-based active learning.}
% \label{algo:al}
% \begin{algorithmic}
% \State \textbf{Input:}
% \State \quad $\mathcal{D}\gets\{1,2,...,n\}$ \quad \% a data set of $n$ instances
% \State \textbf{Initialize:}
% \State \quad $\mathcal{D}_l\gets\{s_1, s_2,...,s_k\}$ \quad\% initial labeled set with $k$ seeds
% \State \quad $\mathcal{D}_u\gets\mathcal{D} \backslash \mathcal{D}_l$ \quad \% initial unlabeled set
% \While {$\mathcal{D}_u\neq\emptyset$}
% \State Select $s^*$ from $\mathcal{D}_u$ \quad\% according to a query strategy
% \State Query the label $y_{s^*}$ for the selected instance $s^*$
% \State $\mathcal{D}_l\gets\mathcal{D}_l\cup\{s^*\}$
% \State $\mathcal{D}_u\gets \mathcal{D}_u\backslash\{s^*\}$
% \EndWhile
% \end{algorithmic}
% \end{algorithm}

\subsection{Query Strategies}
The key component of active learning is the design of an effective criterion for selecting the most ``valuable'' instance to query, which is often referred to as \emph{query strategy}.
We use $s^*$ to refer to the selected instance by the strategy.
In general, different strategies follow a greedy framework:
\begin{equation}\label{eq:general}
\vspace{-0.1cm}
s^*=\argmax_{s\in D_u} \min_{y\in \{-1,1\}} f(s; y, D_l),
\vspace{-0.1cm}
\end{equation}
where $f(s; y, \mathcal D_l)\in \mathbb R$
is a scoring function to measure the risks of choosing $y$
as the label for $\mathbf x_s\in D_u$ given an existing labeled set $D_l$.

We investigate two commonly used query strategies: uncertainty sampling~\cite{lewis1994heterogeneous} and query-by-committee~\cite{seung1992query}. We show that under the binary classification setting, they can all be reformulated as Eq.~\eqref{eq:general}.

\noindent \textbf{Uncertainty Sampling} selects the instance which it is least certain how to label. We choose to study one popular uncertainty-based sampling variant, the \emph{least confident}. Subject to Eq.~\eqref{eq:general}, the resulting approach is to let
\begin{equation}
\vspace{-0.1cm}
f(s; y, \mathcal D_l)= 1 - P_{\Delta(\mathcal D_l)}(y_s = y|\mathbf{x}_s),
\vspace{-0.1cm}
\end{equation}
where $P_{\Delta(\mathcal D_l)}(y_s=y|\mathbf{x}_s)$ is a conditional probability which is estimated from a probabilistic classification model $\Delta$ trained on
$\{(\mathbf x_i, y_i) \mid \forall i\in\mathcal D_l\}$.

\noindent \textbf{Query-By-Committee} maintains a committee of models trained on labeled data, $\mathcal{C}(\mathcal D_l)=\{g^{(1)},...,g^{(C)}\}$. It aims to reduce the size of version space. Specifically, it selects the unlabeled instance about which committee members disagree the most based on their predictions. Subject to Eq.~\eqref{eq:general}, the resulting approach is to let
% \emph{vote entropy}:
\begin{equation}
\vspace{-0.1cm}
f(s; y, \mathcal D_l)=\sum\nolimits_{k=1}^C \mathbf 1[y\neq g^{(k)}(\mathbf x_s)],
\vspace{-0.1cm}
\end{equation}
where $g^{(k)}(\mathbf x_s)\in \{-1,1\}$ is the predicted label of $\mathbf x_s$ using the classifier $g^{(k)}$.

Our paper will start from generalizing Eq.~\eqref{eq:general}
and show 
% in Section~\ref{sec:rquery} 
that it is possible to
extend the two popular query strategies for considering relational data as a strict order.

\section{Active Learning of a Strict Order}\label{sec:method}
Given $G$ a strict order of $V$, consider a set of data $\mathcal D\subseteq V\times V$, where $(a,a)\not\in \mathcal D,\forall a\in V$. Similar to the pool-based active learning, one needs to maintain a labeled set $D_l$ and an unlabeled set $D_u$. We require that $\mathcal D\subseteq \mathcal D_l\cup \mathcal D_u$and $\mathcal D_l\cap \mathcal D_u=\varnothing$.
Given a feature extractor $\mathcal F: V\times V \mapsto \mathbb R^d$, we can build a vector dataset $\{\mathbf x_{(a,b)} = \mathcal F(a,b)\in \mathbb R^d \mid (a,b)\in \mathcal D\}$. Let $y_{(a,b)}= -1 + 2\cdot \mathbf 1[(a,b)\in G]\in\{-1,1\}$ be the ground-truth label for each $(a,b)\in V\times V$. Active learning aims to query $Q$ a subset from $\mathcal D$ under limited budget and construct a label set $\mathcal D_l$ from $Q$, in order to train a good classifier $h$ on $\mathcal D_l\cap \mathcal D$ such that it predicts accurately whether or not an unlabeled pair $(a,b)\in G$ by $h(\mathcal F(a,b))\in \{-1,1\}$.

Active learning of strict orders differs from the traditional active learning in two unique aspects:
(i) By querying the label of a single unlabeled instance, one may obtain a set of labeled examples, with the help of strict orders' properties;
(ii) The relational information of strict orders could also be utilized by query strategies.
% In Section~\ref{sec:rel-al} and Section~\ref{sec:rquery}, 
We will present our efforts towards incorporating the above two aspects into active learning of a strict order.

\subsection{Basic Relational Reasoning in Active Learning}\label{sec:rel-al}
A basic extension from standard active learning to one under the strict order setting is to apply relational reasoning when both updating $\mathcal D_l$ and predicting labels.
Algorithm~\ref{algo:strict} shows the pseudocode for the pool-based active learning of a strict order.
When updating $\mathcal D_l$ with a new instance $(a,b)\in \mathcal D_u$ whose label $y_{(a,b)}$ is acquired from querying,
one first calculates $\overline{\mathcal D_l'}$, i.e., the closure of $\mathcal D_l\cup \{(a,b)\}$,
using Theorem~\ref{thm:cxy}, and then sets $\mathcal D_l:=\overline{\mathcal D_l'}$ and $\mathcal D_u:=D\backslash \overline{\mathcal D_l'}$ respectively. Therefore, it is possible to augment the labeled set $\mathcal D_l$ with more than one pair at each stage even though only a single instance is queried.
Furthermore, the following corollary shows that given a fixed set of samples to be queried, their querying order does not affect the final labeled set $D_l$ constructed.
\begin{corollary}~\label{cor:consistent}
Given a list of pairs $Q$ of size $m$ whose elements are from $V\times V$, let $i_1,\ldots, i_m$ and $j_1,\ldots,j_m$ be two different permutations of $1,\ldots,m$. Let $I_0=\varnothing$ and $J_0=\varnothing$, and $I_k = \overline{I_{k-1}\cup \{q_{i_k}\}}$, $J_k = \overline{J_{k-1}\cup \{q_{j_k}\}}$ for $k=1,\ldots,m$, where $\overline{\cdot}$ is defined as the closure set under $G$. We have $I_m=J_m$, which is the closure of $\{q_i\in V\times V \mid i=1,\ldots,m\}$.
\end{corollary}

\begin{algorithm}[t]
\caption{Pseudocode for pool-based active learning of a strict order.}
\label{algo:strict}
\begin{algorithmic}
\State \textbf{Input:}
\State \quad $\mathcal{D}\subseteq V\times V$ \quad \% a data set
\State \textbf{Initialize:}
\State \quad $\mathcal{D}_l\gets\{(a_{s_1},b_{s_1}), (a_{s_2},b_{s_2}),...,(a_{s_k},b_{s_k})\}$ \quad\% initial labeled set with $k$ seeds
\State \quad $\mathcal{D}_l\gets \overline{\mathcal{D}_l}$ \quad \% initial closure
\State \quad $\mathcal{D}_u\gets\mathcal{D} \backslash \mathcal{D}_l$ \quad \% initial unlabeled set
\While {$\mathcal{D}_u\neq\emptyset$}
\State Select $(a^*, b^*)$ from $\mathcal{D}_u$ \quad\% according to a query strategy
\State Query the label $y_{(a^*,b^*)}$ for the selected instance $(a^*,b^*)$
\State $\mathcal{D}_l\gets\overline{\mathcal{D}_l\cup\{(a^*,b^*)\}}$
\State $\mathcal{D}_u\gets \mathcal{D}\backslash\mathcal{D}_l$
\EndWhile
\end{algorithmic}
\end{algorithm}

Corollary~\ref{cor:consistent} is a straightforward result from the uniqueness of closure,
which is also verified by our experiments.
The labeled set $\mathcal D_l$ contains two kinds of pairs based on where their labels come from:
The first kind of labels comes directly from queries, and the second kind comes from the relational reasoning
as explained by Theorem~\ref{thm:cxy}. Such an approach has a clear advantage over
standard active learning at the same budget of queries,
because labels of part of the test pairs can be inferred deterministically
and as a result there will be more labeled data
for supervised training.
In our setup of active learning, we train classifiers on $\mathcal D \cap \mathcal D_l$ and use
them for predicting the labels of remaining pairs that are not in $\mathcal D_l$.

\subsection{Query Strategies with Relational Reasoning}\label{sec:rquery}
The relational active learning framework as explained in the previous section however does not consider incorporating relational reasoning in its query strategy. 
We further develop a systematic approach on how to achieve this.

\vspace{0.2cm}
We start from the following formulation:
at each stage, one chooses a pair $(a^\ast, b^\ast)$ to query based on
\small
\begin{eqnarray}\label{eq:general2}
\vspace{-0.1cm}
\!\!\!\!\!\!(a^\ast, b^\ast)\!\!\!&=&\!\!\!\argmax_{(a,b)\in \mathcal D_u} \min_{y\in \{-1,1\}} F(\mathcal S(y_{(a,b)}= y), \mathcal D_l), \\
&& \!\!\!\!\!\!\mathcal S(y_{(a,b)}= y) = (\overline{\mathcal D_l\cup \{(a,b)\}} \backslash \mathcal D_l) \cap \mathcal D.
\vspace{-0.1cm}
\end{eqnarray}
\normalsize
Again, $F$ is the scoring function.
$\mathcal S(y_{(a,b)}= y)$ is the set of pairs in $\mathcal D$ whose labels,
originally unknown ($\not\in \mathcal D_l$), can now be
inferred by assuming $y_{(a,b)}=y$ using Theorem~\ref{thm:cxy}.
For each $(u,v)\in \mathcal S(y_{(a,b)}= y)$,
its inferred label is denoted as $\hat y_{(u,v)}$ in the sequel.
One can see that this formulation is a generalization of Eq.~\eqref{eq:general}.
We now proceed to develop extensions for the two query strategies
% as discussed in Section~\ref{sec:al}
to model the dependencies between pairs imposed by the rule of a strict order.
Following the same notations as described in
Section~\ref{sec:al} 
with the only difference that
the numbering index is replaced by the pairwise index, we propose two query strategies tailored to strict orders.

\vspace{0.2cm}
\noindent \textit{Uncertainty Sampling with Reasoning.} With relational reasoning, one not only
can reduce the uncertainty of the queried pair $(a,b)$ but also may reduce that of other pairs deduced by assuming $y_{(a,b)}\!\!=\!\!y$. The modified scoring function reads:
\small
\begin{multline}
\vspace{-0.1cm}
\!\!\!\!\!\!F(\mathcal S(y_{(a,b)}= y), \mathcal D_l) = \\
\!\!\!\!\!\!\sum_{(u,v)\in \mathcal S(y_{(a,b)}= y)} \!\!\!\!1 - P_{\Delta(\mathcal D_l\cap \mathcal D)}(y_{(u,v)} = \hat{y}_{(u,v)} | \mathbf x_{(u,v)}).
\vspace{-0.1cm}
\end{multline}
\normalsize

\noindent \textit{Query-by-Committee with Reasoning.} Likewise, one also has the extension
for QBC, where $\{g^{(k)}\}_{k=1}^C$ is a committee of classifiers trained on bagging
samples of $\mathcal D_l\cap \mathcal D$,
\small
\begin{multline}
\vspace{-0.1cm}
\!\!\!\!\!\!F(\mathcal S(y_{(a,b)}= y), \mathcal D_l) = \\
\!\!\!\!\!\!\sum_{(u,v)\in \mathcal S(y_{(a,b)}= y)} \sum\nolimits_{k=1}^C \mathbf 1 (\hat{y}_{(u,v)} \neq g^{(k)}(\mathbf x_{(u,v)})).
\vspace{-0.1cm}
\end{multline}
\normalsize

\subsection{Bounds on the Number of Queries}
Note that any strict order $G$ can be described as a directed acyclic graph (DAG). We show the lower and upper bounds on the number of queries that are needed to learn a consistent classifier for $G$.
\begin{theorem}\label{thm:bound}
Given a strict order G of V, let A be a consistent learner that makes $m$ queries to the oracle before termination, then
\begin{equation}
\vspace{-0.2cm}
|\underline{G}|\leq m \leq |\overline{G}|
\vspace{-0.1cm}
\end{equation}
where \underline{G} is the transitive reduction~\cite{aho1972transitive} of G and $\overline{G}$ is the closure (Def.~\ref{def:closure}) of G.
\end{theorem}
The above lower and upper bounds are tight in the sense that there exist DAGs $G_1$ and $G_2$, such that $|\underline{G_1}|=|G_1|$ and $|\overline{G_2}|=|G_2|$. With the power of the active learning paradigm, we want to empirically show that the number of queries needed is much smaller.

\section{Experiments}
\label{sec:exp}
For evaluation, we apply the proposed active learning algorithms to \emph{concept prerequisite learning problem}~\cite{DBLP:conf/emnlp/LiangWHG15,pan2017prerequisite}.
Given a pair of concepts (A, B), we predict whether or not A is a prerequisite of B, which is a binary classification problem. Here, cases where B is a prerequisite of A and where no prerequisite relation exists are both considered negative.
% Prerequisite relation, as a strict partial order, plays an important role for many educational applications. Despite there have been earlier work on automatic discovery of concept prerequisite relations~\cite{talukdar2012crowdsourced,vuong2010method},
% the lack of labeled concept pairs has made scalable data driven solutions impractical.
% Applying active learning to tackle this challenge is a new opportunity.
% In addition, previous work have not exploited the relational structure of concept prerequisites as strict partial orders.

\subsection{Dataset}
We use the Wiki concept map\footnote{Each concept corresponds to an English Wiki article.}
dataset from~\cite{wang2016using} which is collected from textbooks on different educational domains.
For each domain, the dataset consists of prerequisite pairs in the concept map.
% In the preprocessing stage,
% we validate whether each of the prerequisite relations in the dataset satisfies the required properties of a strict partial order and ask domain experts to manually correct their labels if needed. We also expand the dataset by using the irreflexive and transitive properties: (i) add (B, A) as a negative sample if (A, B) is a positive sample; (ii) add (A, C) as a positive sample if both (A, B) and (B, C) are positive samples.
Table~\ref{tab:data} summarizes the statistics of the our final processed dataset.

\begin{table}[tp]
\centering
\scalebox{0.8}{
\begin{tabular}{lccc}
\toprule
\textbf{Domain} & \textbf{\# Concepts} & \textbf{\# Pairs} & \textbf{\# Prerequisites}\\
\midrule
Data Mining & 120 & 826& 292\\
Geometry & 89&1681& 524\\
Physics & 153&1962& 487\\
Precalculus &224 &2060& 699\\
\bottomrule
%\vspace{-0.5cm}
\end{tabular}
}
\vspace{-0.2cm}
\caption{Dataset statistics.}
\label{tab:data}
\vspace{-0.5cm}
\end{table}

\subsection{Features}
For each concept pair $(A, B)$, we calculate two types of features following the popular practice of information retrieval and natural language processing: graph-based features and text-based features. Please refer to Table~\ref{tab:graph} for detailed description.
Note we trained a topic model~\cite{blei2003latent} on the Wiki corpus. We also trained a Word2Vec~\cite{mikolov2013distributed} model on the same corpus with each concept treated as an individual token.

\begin{table}[t]
\centering
\scalebox{0.68}{
\begin{tabular}{lp{7.5cm}}
\toprule
\textbf{Feature} & \textbf{Description}\\
\midrule
In/Out Degree & The in/out degree of A/B.\\
Common Neighbors & \# common neighbors of A and B.\\
\# Links & \# times A/B links to B/A.\\
Link Proportion & The proportion of pages that link to A/B also link to B/A.\\
NGD & The Normalized Google Distance between A and B~\cite{witten2008effective}.\\
PMI & The Pointwise Mutual Information relatedness between the incoming links of A and B~\cite{ratinov2011local}.\\
RefD & A metric to measure how differently A and B's related concepts refer to each other~\cite{DBLP:conf/emnlp/LiangWHG15}.\\
HITS & The difference between A and B's hub/authority scores.~\cite{kleinberg1999authoritative}\\
\midrule
1st Sent & Whether A/B is in the first sentence of B/A.\\
In Title & Whether A appears in B's title.\\
Title Jaccard & The Jaccard similarity between A and B's titles.\\
Length & \# words of A/B's content.\\
Mention & \# times A/B are mentioned in the content of B/A.\\
NP & \# noun phrases in A/B's content; \# common noun phrases.\\
Tf-idf Sim & The cosine similarity between Tf-idf vectors for A and B's first paragraphs.\\
Word2Vec Sim& The cosine similarity between vectors of A and B trained by Word2Vec.\\
LDA Entropy & The Shannon entropy of the LDA vector of A/B.\\
LDA Cross Entropy & The cross entropy between the LDA vector of A/B and B/A~\cite{DBLP:conf/acl/GordonZGNB16}.\\
\bottomrule
\end{tabular}
}
\vspace{-0.3cm}
\caption{Feature description. Top: graph-based features. Bottom: text-based features.}
\label{tab:graph}
\vspace{-0.3cm}
\end{table}

\subsection{Experiment Settings}
\label{sec:expsetting}
We follow the typical evaluation protocol of pool-based active learning.
We first randomly split a dataset into a training set $\mathcal D$ and a test set $\mathcal D_{test}$ with a ratio of 2:1.
Then we randomly select 20 samples from the training set as the initial query set $Q$ and
compute its closure $\mathcal D_l$. Meanwhile, we set $\mathcal D_u=\mathcal D\backslash \mathcal D_l$.
In each iteration, we pick an unlabeled instance from $\mathcal D_u$
to query for its label, update the label set $\mathcal D_l$, and re-train a classification model
on the updated $\mathcal D_l\cap \mathcal D$.
The re-trained classification model is then evaluated on $\mathcal D_{test}$. In all experiments, we use a random forests classifier~\cite{breiman2001random} with 200 trees as the classification model.
We use Area under the ROC curve (AUC) as the evaluation metric.
Taking into account the effects of randomness subject to different initializations, we continue the above experimental process for each method repeatedly with 300 preselected distinct random seeds. Their average scores and confidence intervals ($\alpha =0.05$) are reported. We compare four query strategies:
\begin{itemize}[leftmargin=*]
\item Random: randomly select an instance to query.
\item LC: least confident sampling, a widely used uncertainty sampling variant. We use logistic regression to estimate posterior probabilities.
\item QBC: query-by-committee algorithm. We apply query-by-bagging~\cite{mamitsuka1998query} and use a committee of three decision trees. %We choose the decision tree to test the performance when using a non-probabilistic model.
% \item QUIRE: a strategy for querying informative and representative examples (as discussed in Section~\ref{sec:al}). We follow the authors' experimental approach to use an RBF kernel and set the parameter $\lambda=1$.
\item CNT: a simple baseline query strategy designed to greedily select an instance whose label can potentially infer the most number of unlabeled instances. Following the previous notations, the scoring function for CNT is
\[F(\mathcal S(y_{(a,b)}= y), \mathcal D_l) =|S(y_{(a,b)}= y)|\]
which is solely based on logical reasoning.
\end{itemize}

\begin{table}[tp]
\centering
\scalebox{0.7}{
\begin{tabular}{lp{2cm}p{2cm}p{2cm}}
\toprule
\textbf{Method} & \textbf{Use reasoning when updating $\mathcal D_l$}  & \textbf{Use reasoning to select the instance to query} & \textbf{Use learning to select the instance to query}\\
\midrule
Random & \xmark & \xmark & \xmark \\
LC, QBC & \xmark & \xmark & \cmark \\
Random-R & \cmark & \xmark & \xmark\\
LC-R, QBC-R & \cmark & \xmark & \cmark\\
CNT &\cmark & \cmark & \xmark\\
LC-R+, QBC-R+ & \cmark & \cmark & \cmark\\
\bottomrule
\end{tabular}
}
\vspace{-0.3cm}
\caption{Summary of compared query strategies.}
\label{tab:qs}
\vspace{-0.4cm}
\end{table}

For experiments, we test each query strategy under three settings:
(i) Traditional active learning where no relational information is considered. Query strategies under this setting are denoted as Random, LC, and QBC.
(ii) Relational active learning where relation reasoning is applied to updating $\mathcal D_l$ and predicting labels of $\mathcal D_{test}$.
% , as described in Section~\ref{sec:rel-al}.
Query strategies under this setting are denoted as Random-R, LC-R, and QBC-R.
(iii) Besides being applied to updating $\mathcal D_l$, relational reasoning is also incorporated in the query strategies.
% , as described in Section~\ref{sec:rquery}. 
Query strategies under this setting are the baseline method CNT and our proposed extensions of LC and QBC for strict partial orders, denoted as LC-R+ and QBC-R+, respectively.
Table~\ref{tab:qs} summarizes the query strategies studied in the experiments.

\subsection{Experiment Results}
%In the experiments, we compare different active learning approaches in terms of both effectiveness and efficiency.

\subsubsection{Effectiveness Study}
\begin{figure*}[tp]
\centering
\subfigure[Data Mining - LC]{
\includegraphics[width=\fw\textwidth]{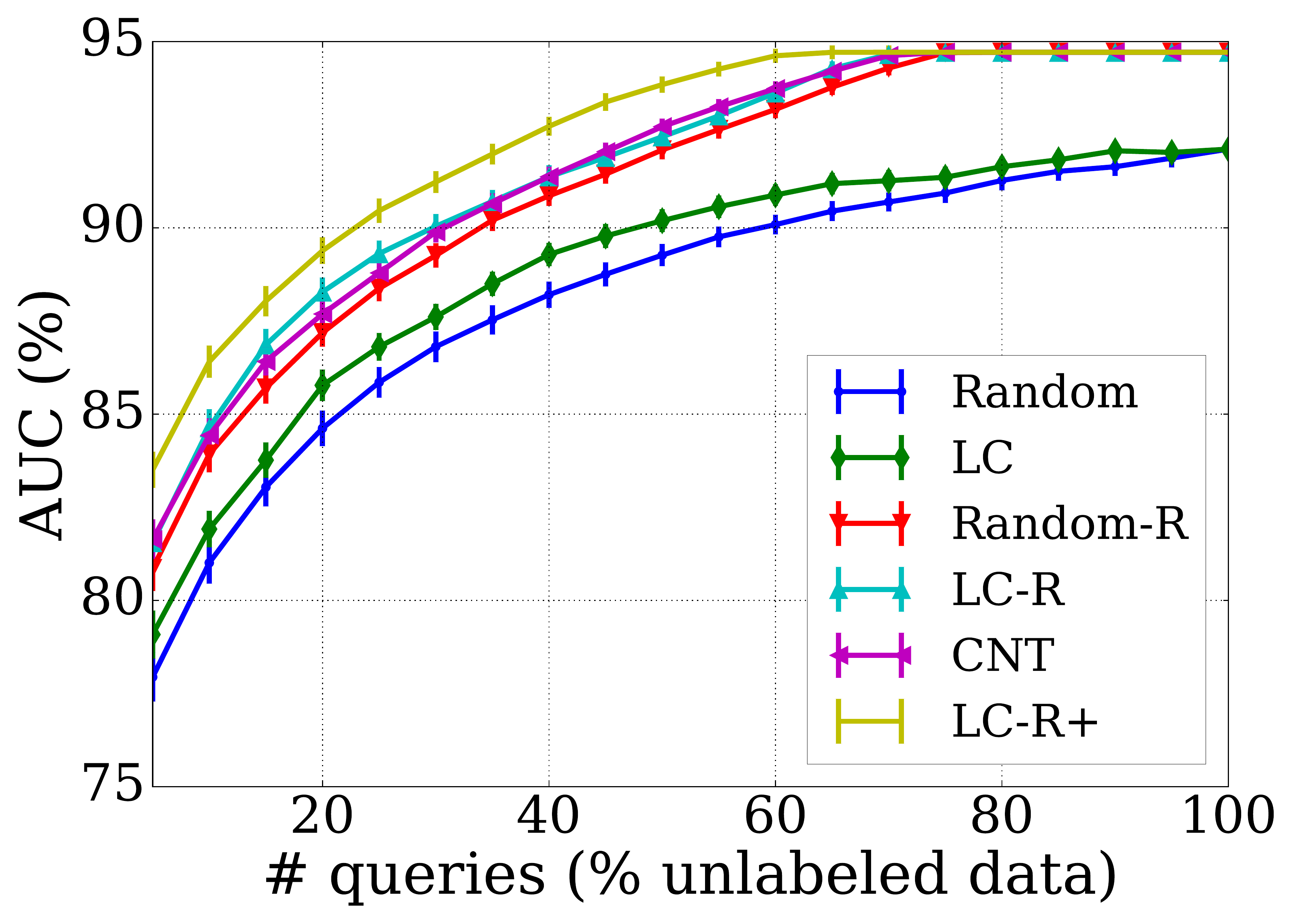}
}
\subfigure[Geometry - LC]{
\includegraphics[width=\fw\textwidth]{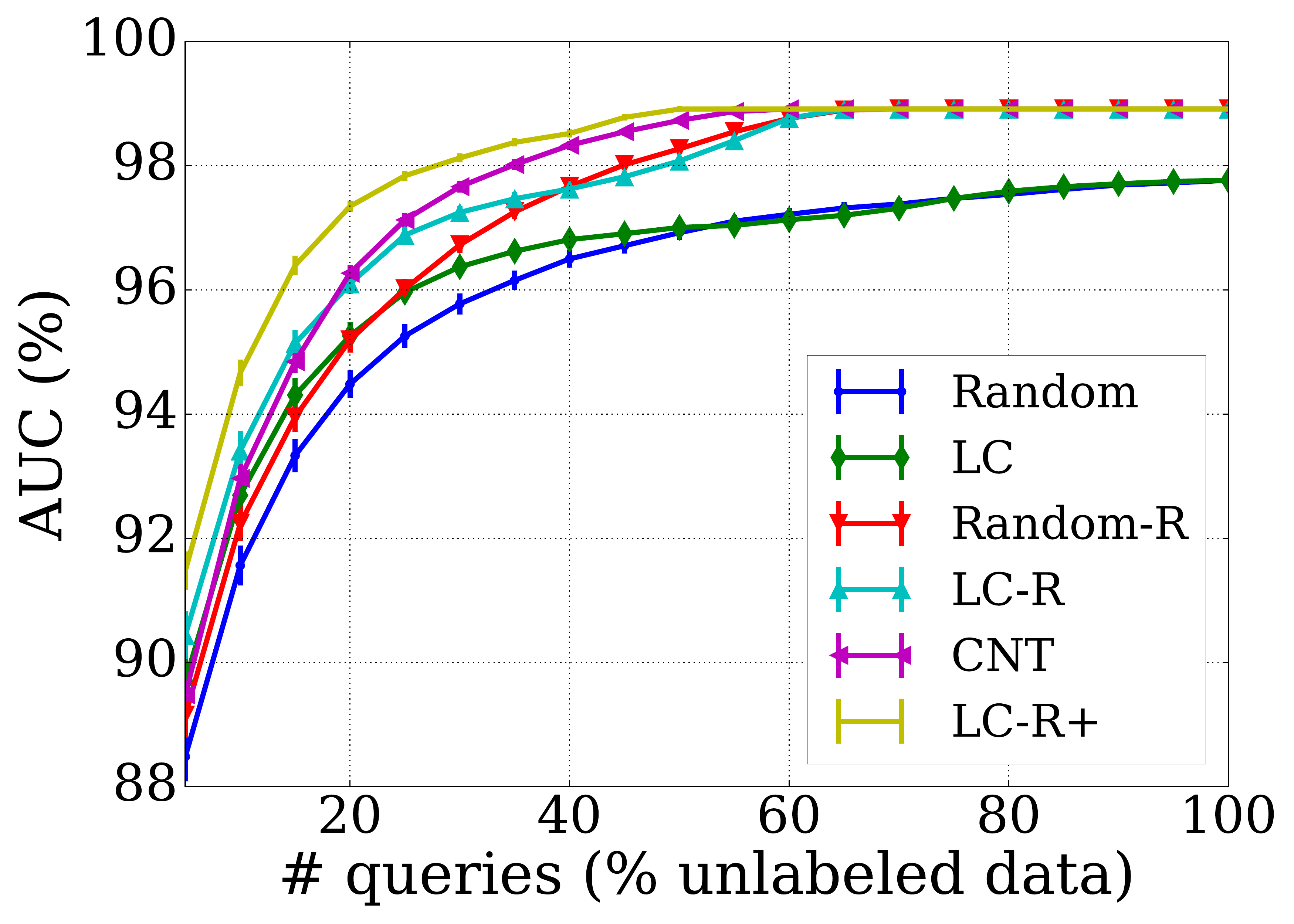}
}
\subfigure[Physics - LC]{
\includegraphics[width=\fw\textwidth]{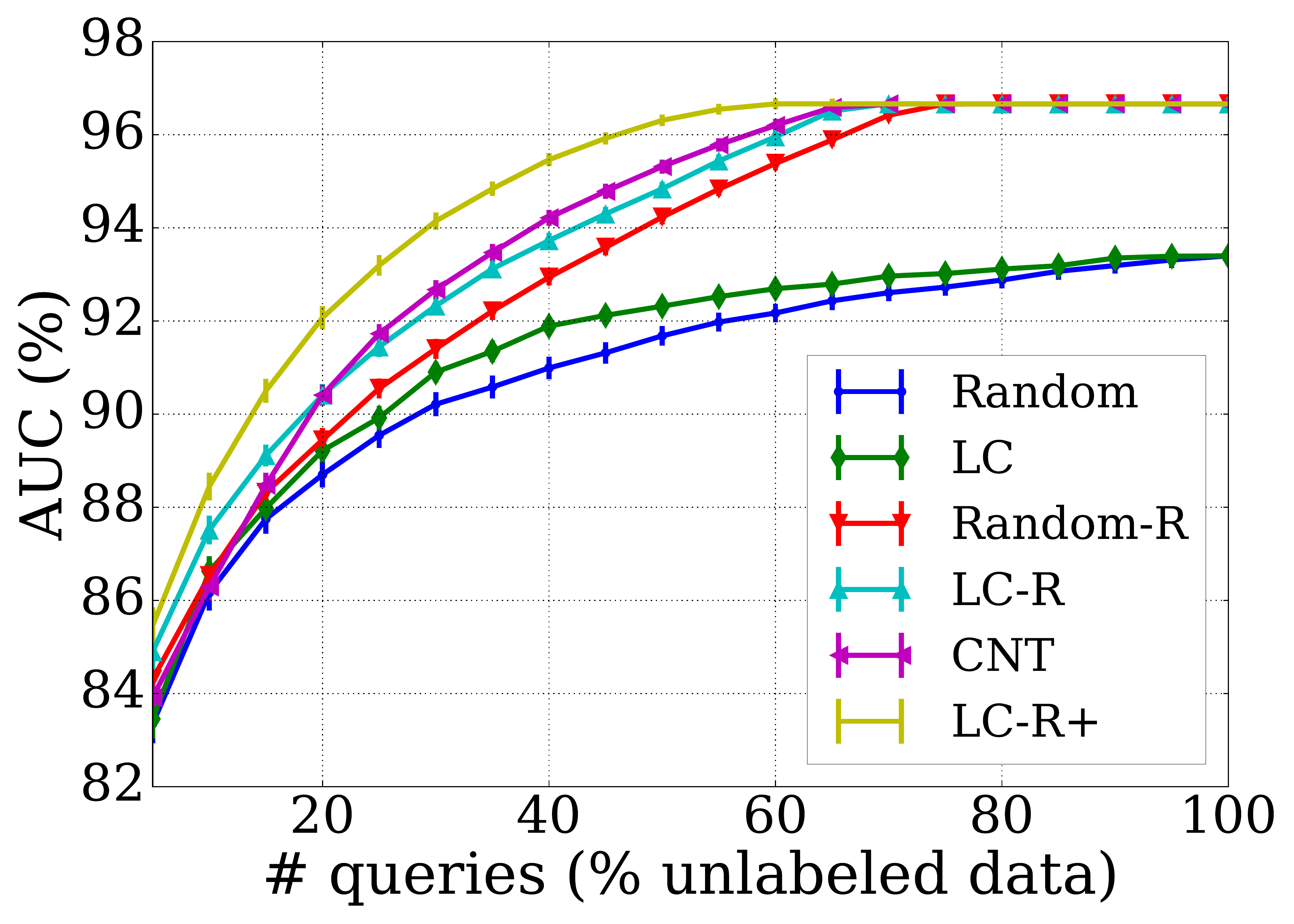}
}
\subfigure[Precalculus - LC]{
\includegraphics[width=\fw\textwidth]{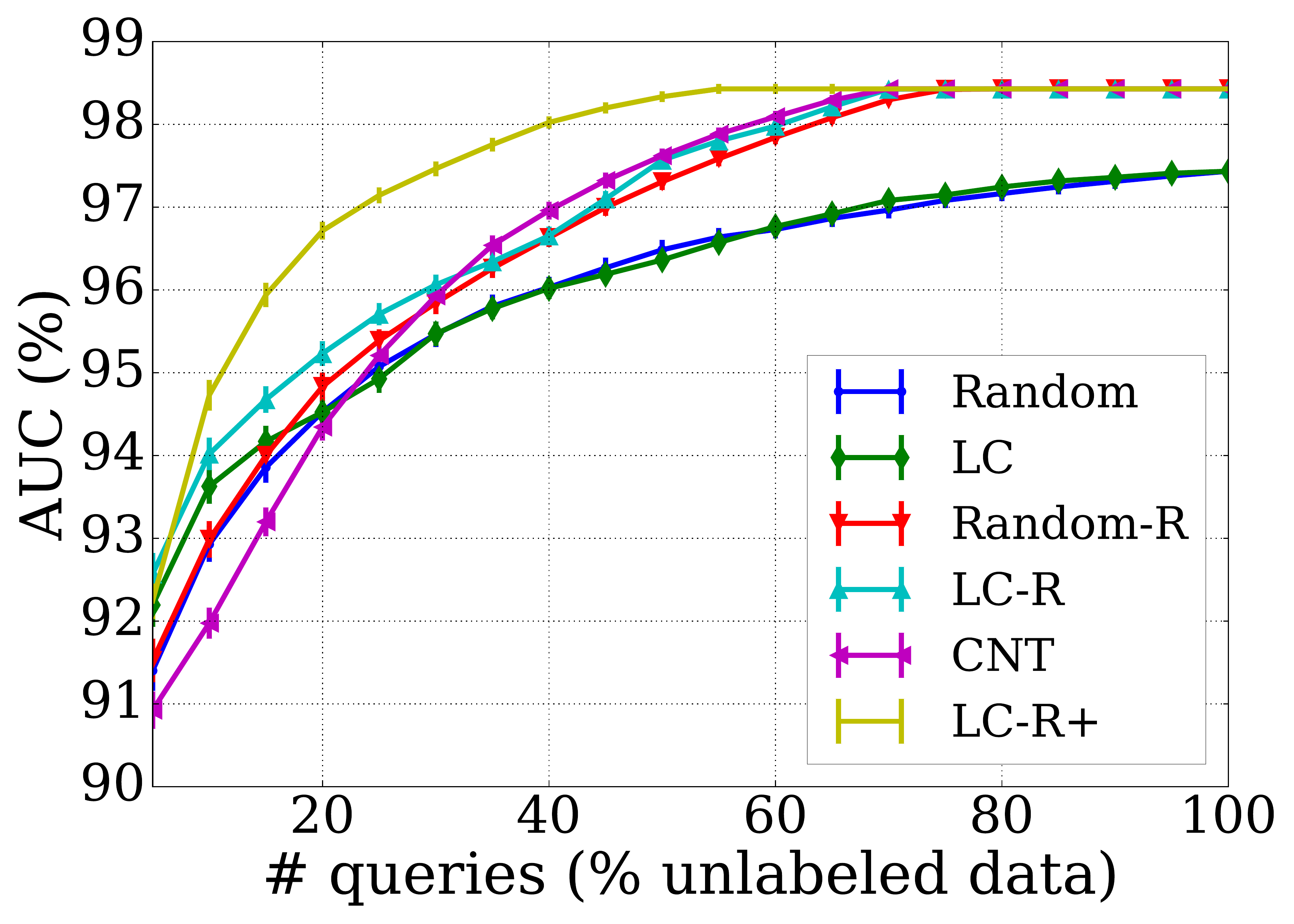}
}
\\
\subfigure[Data Mining - QBC]{
\includegraphics[width=\fw\textwidth]{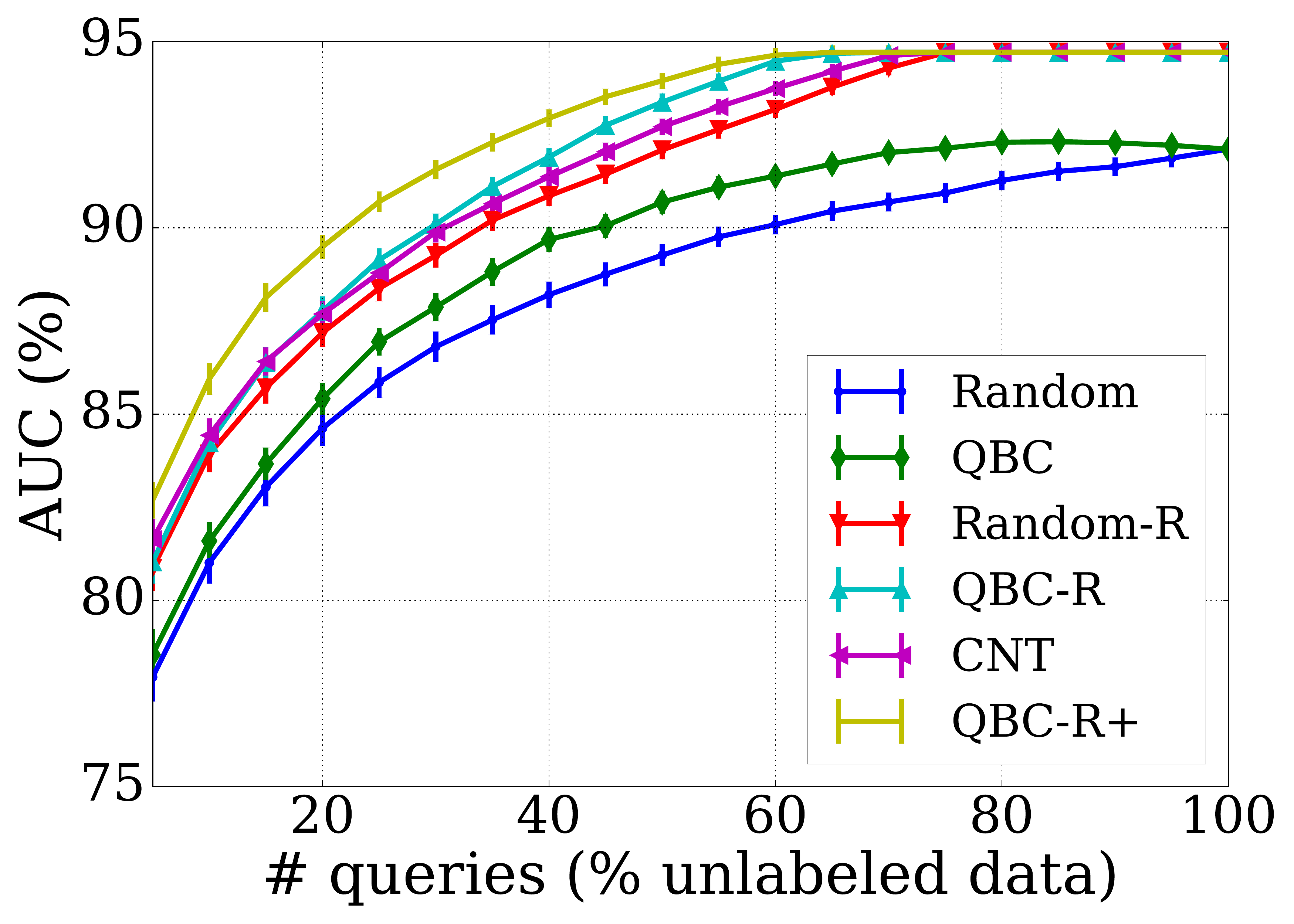}
}
\subfigure[Geometry -QBC]{
\includegraphics[width=\fw\textwidth]{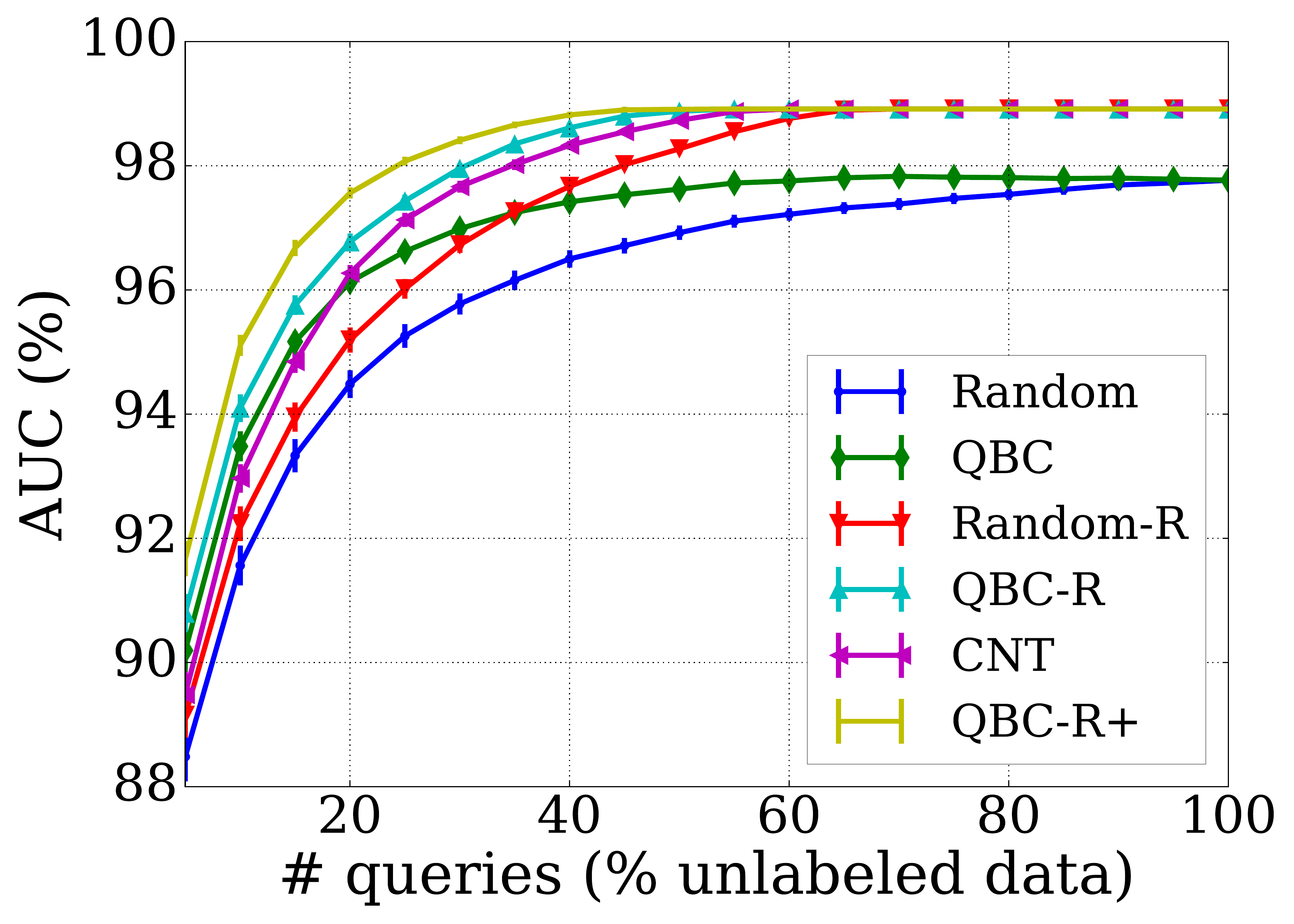}
}
\subfigure[Physics - QBC]{
\includegraphics[width=\fw\textwidth]{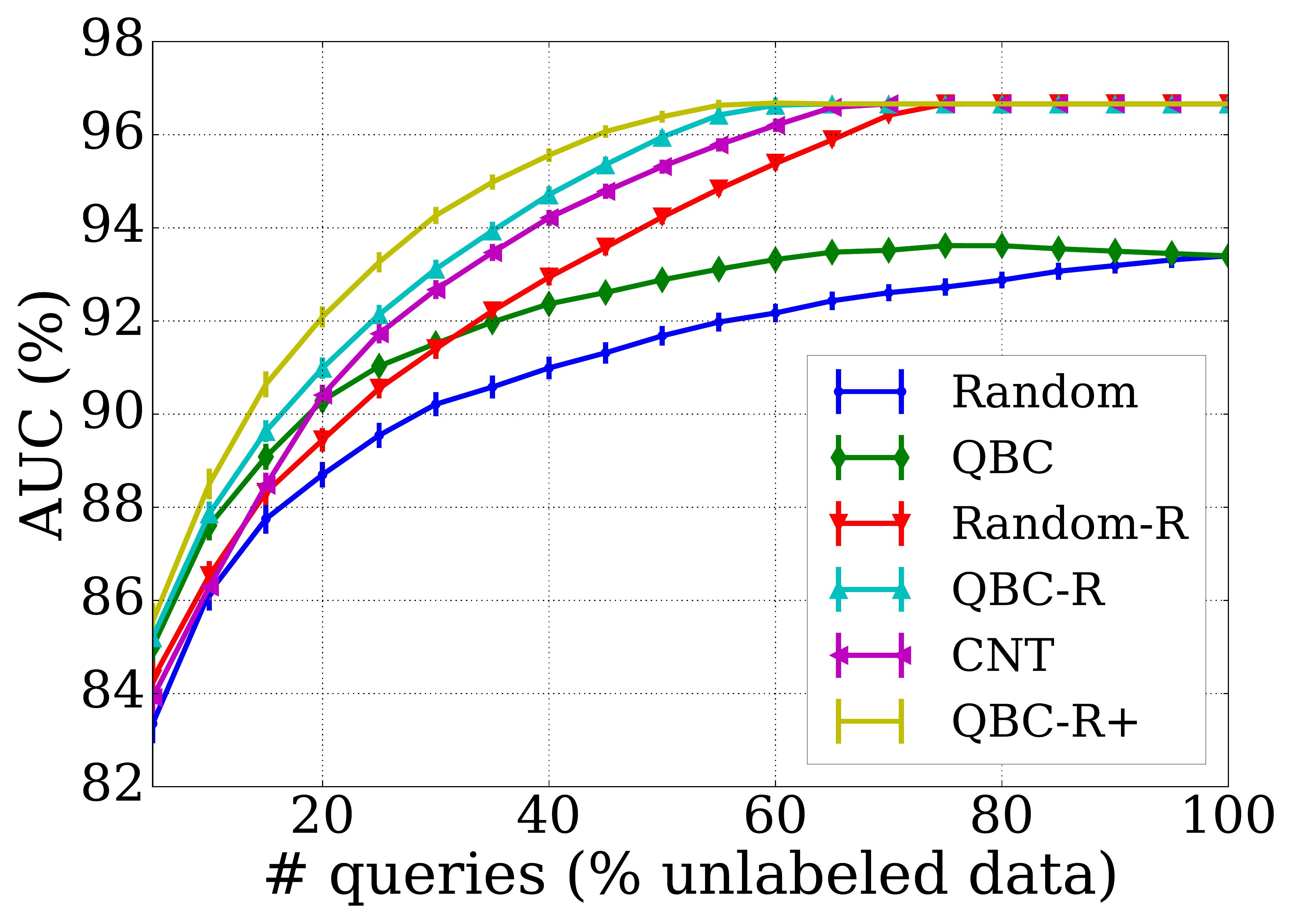}
}
\subfigure[Precalculus - QBC]{
\includegraphics[width=\fw\textwidth]{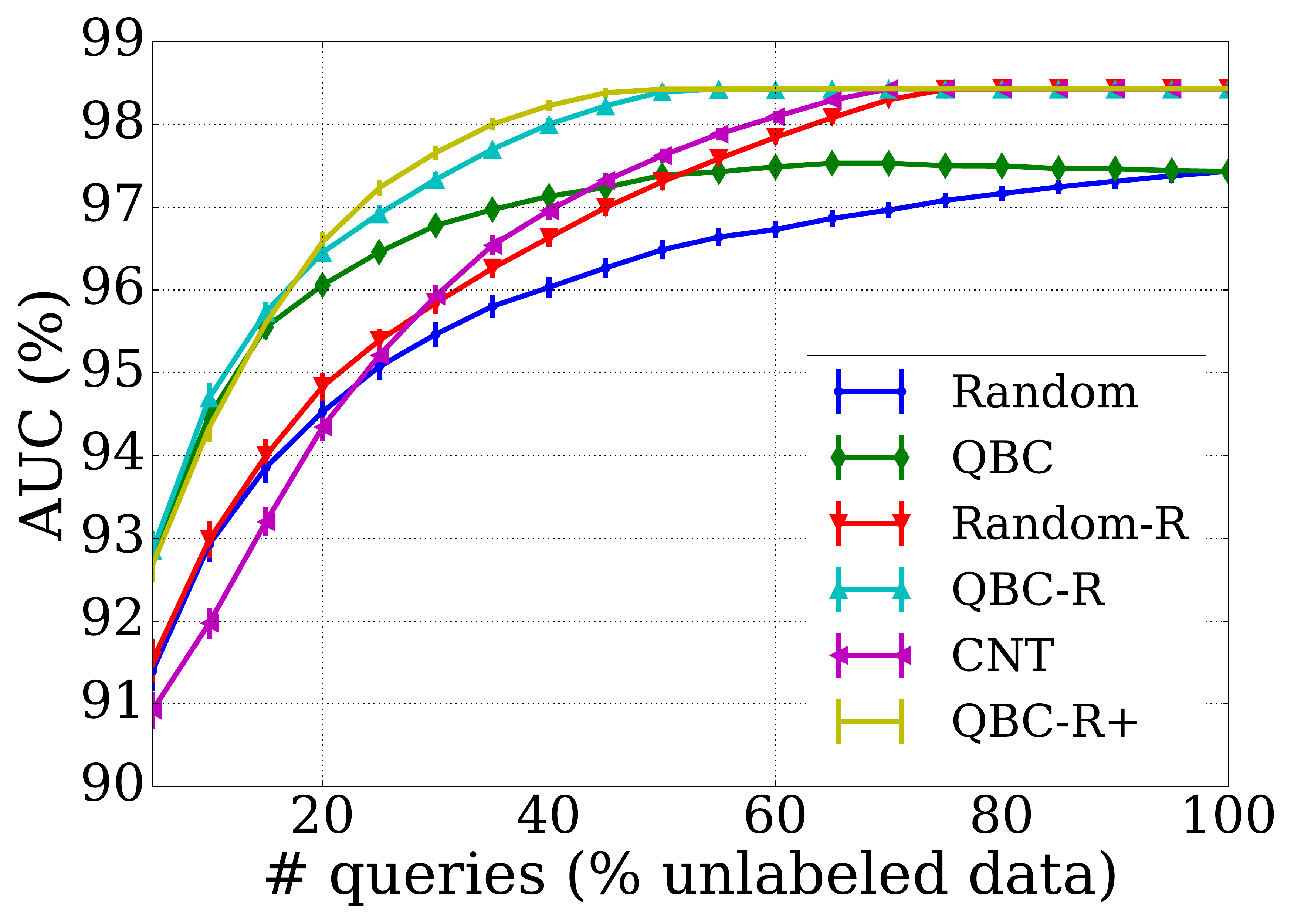}
}
\vspace{-0.3cm}
\caption{Comparison of different query strategies' AUC scores for concept prerequisite learning.}
\label{fig:results}
\vspace{-0.3cm}
\end{figure*}

\begin{figure*}[t]
\centering
\subfigure[Data Mining - LC]{
\includegraphics[width=\fw\textwidth]{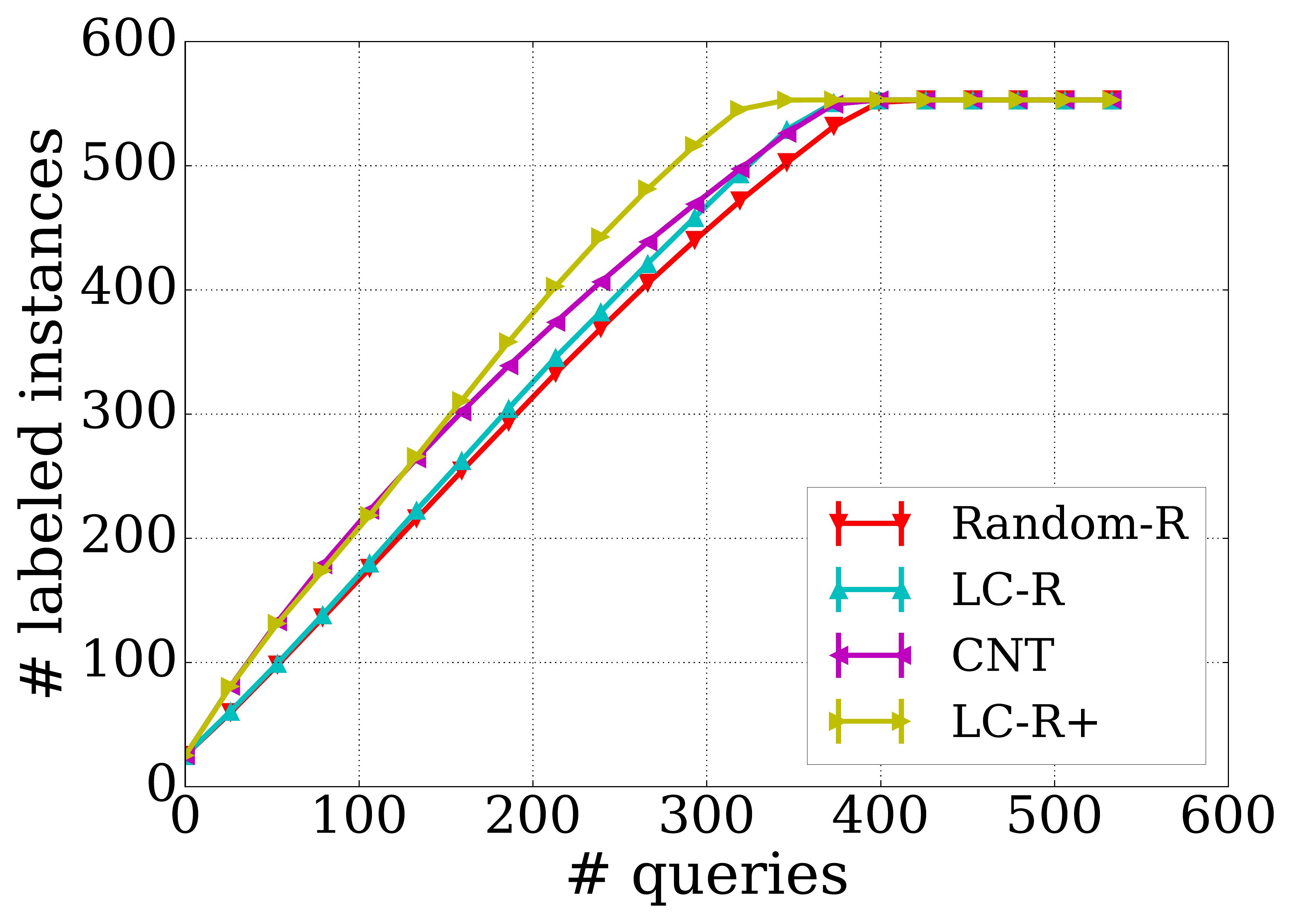}
}
\subfigure[Geometry - LC]{
\includegraphics[width=\fw\textwidth]{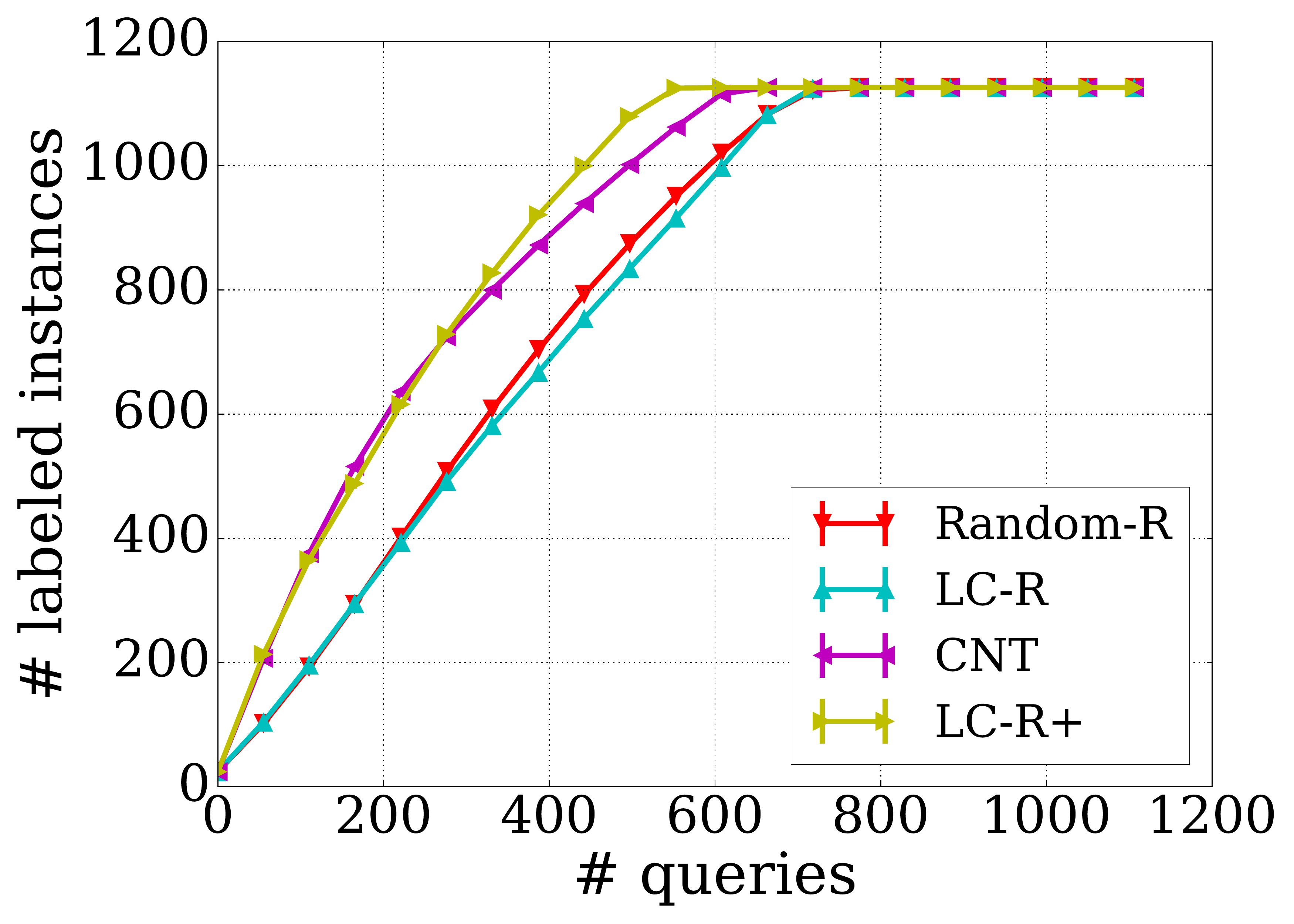}
}
\subfigure[Physics - LC]{
\includegraphics[width=\fw\textwidth]{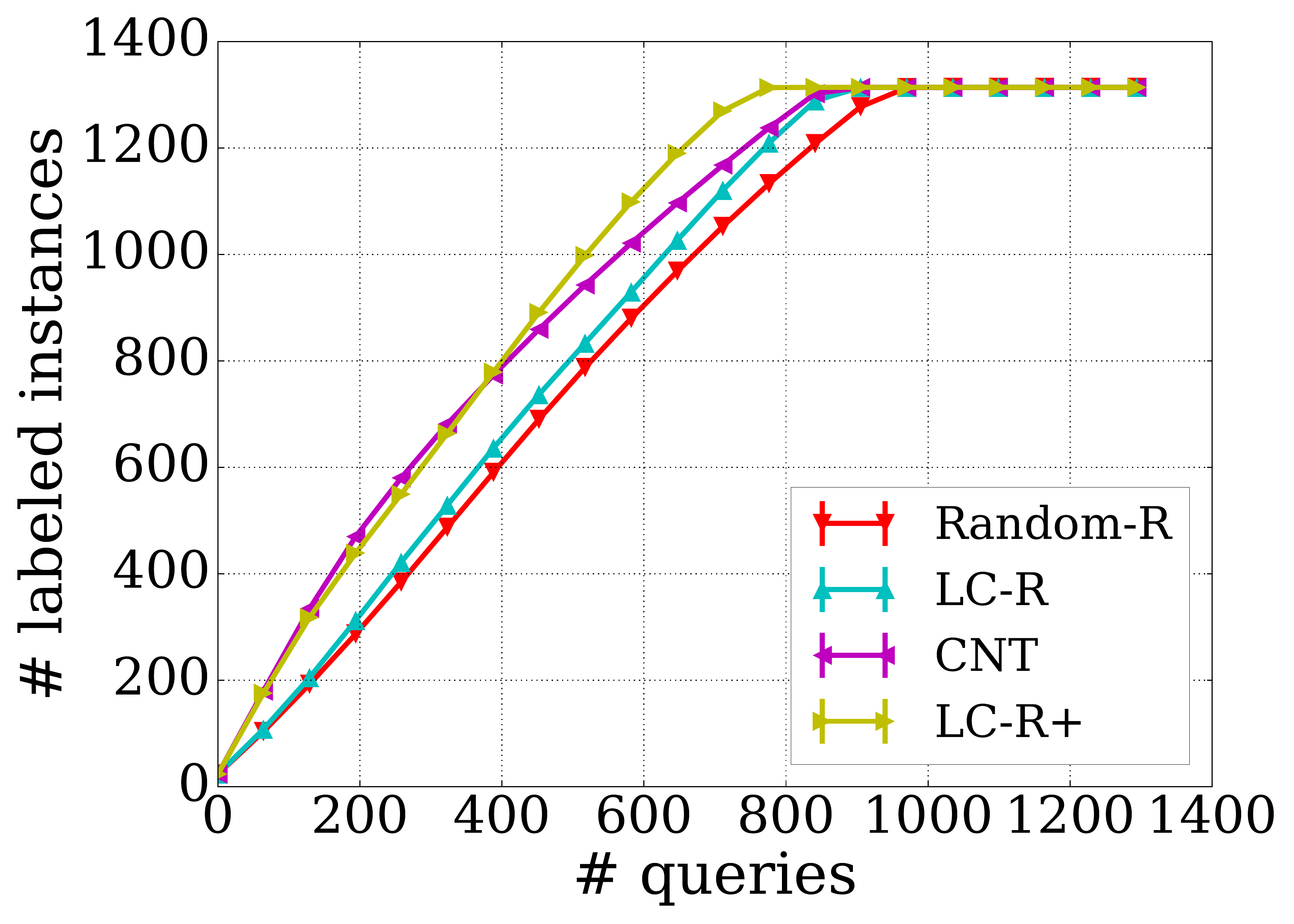}
}
\subfigure[Precalculus - LC]{
\includegraphics[width=\fw\textwidth]{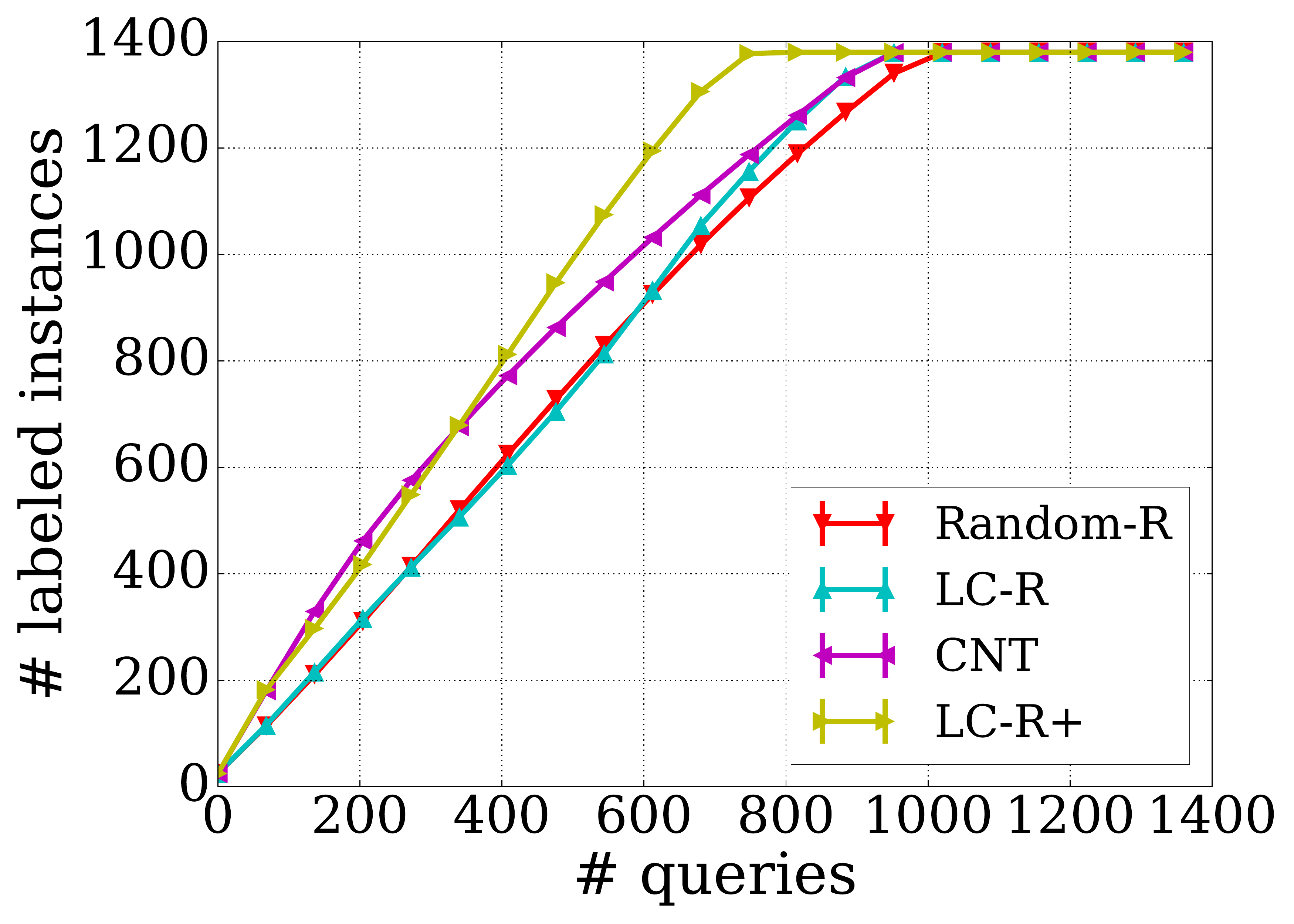}
}
\\
\subfigure[Data Mining - QBC]{
\includegraphics[width=\fw\textwidth]{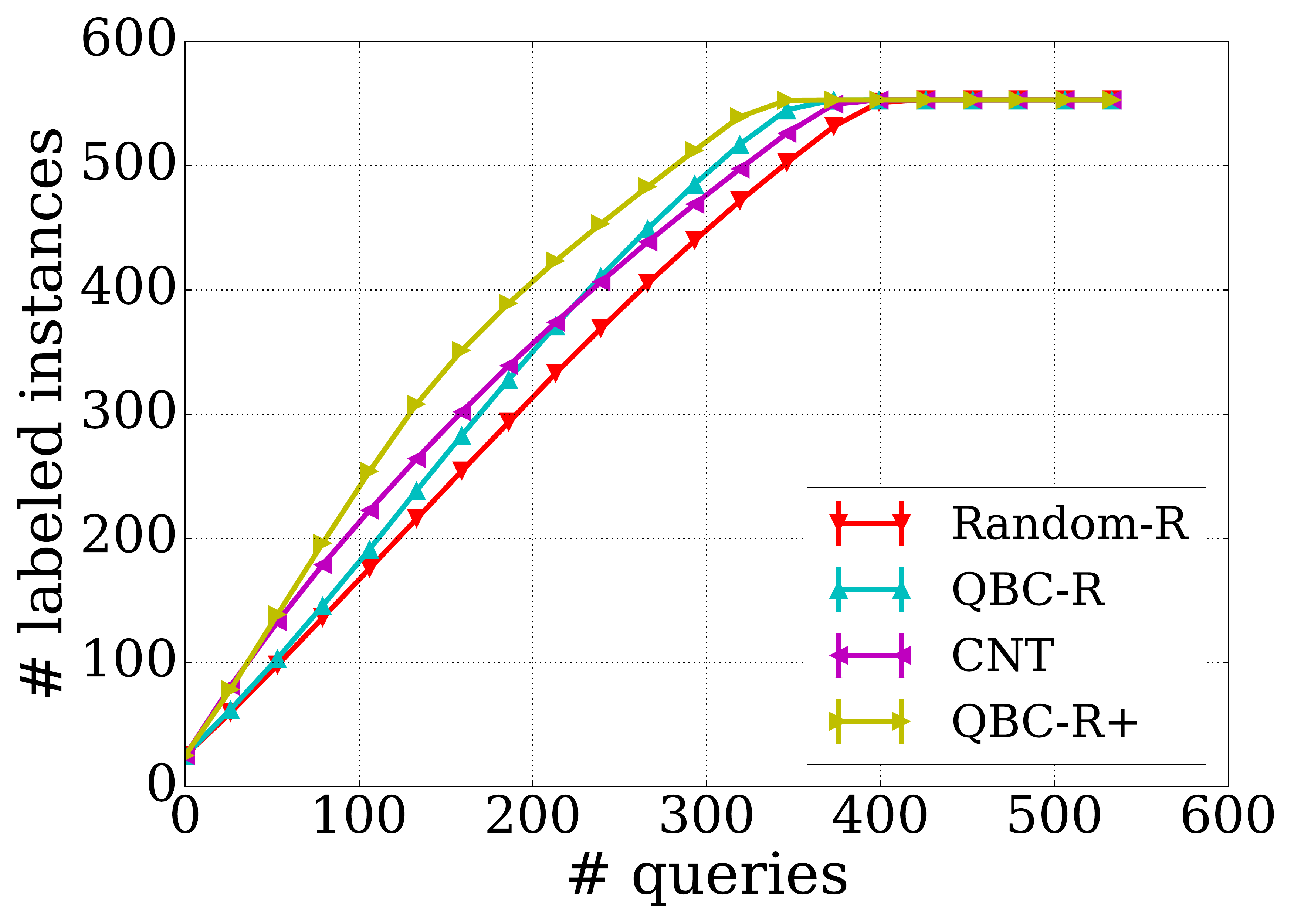}
}
\subfigure[Geometry - QBC]{
\includegraphics[width=\fw\textwidth]{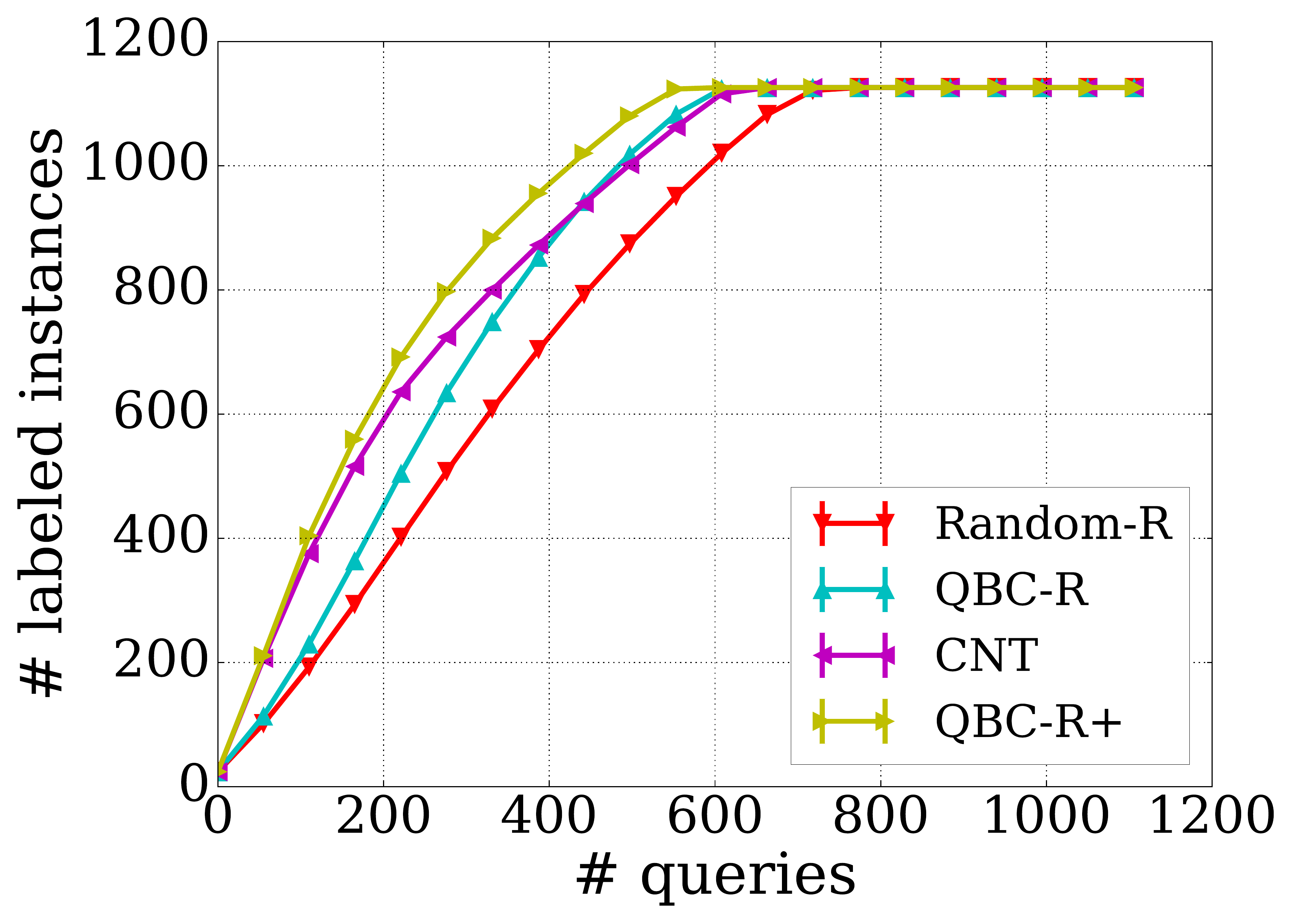}
}
\subfigure[Physics - QBC]{
\includegraphics[width=\fw\textwidth]{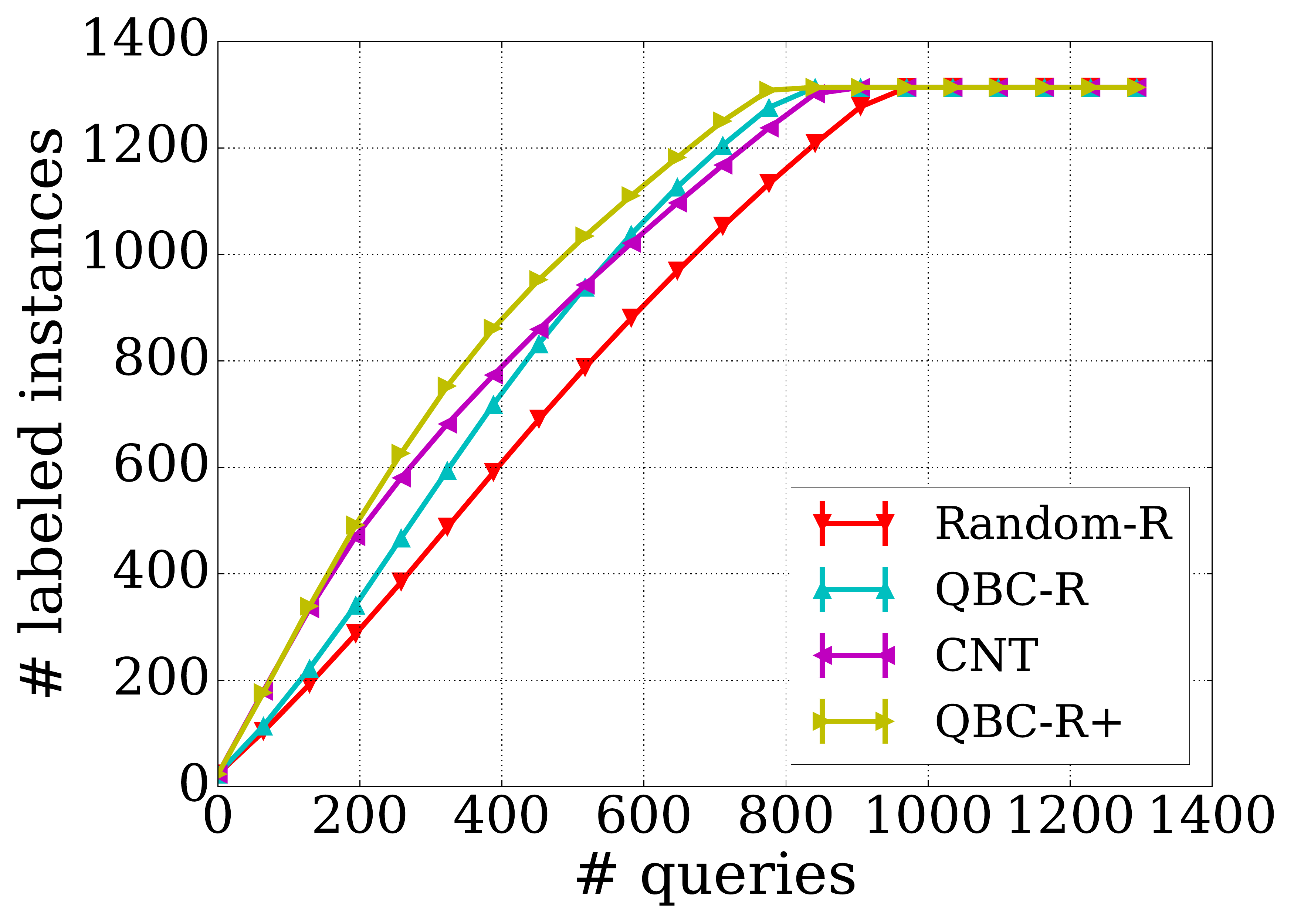}
}
\subfigure[Precalculus - QBC]{
\includegraphics[width=\fw\textwidth]{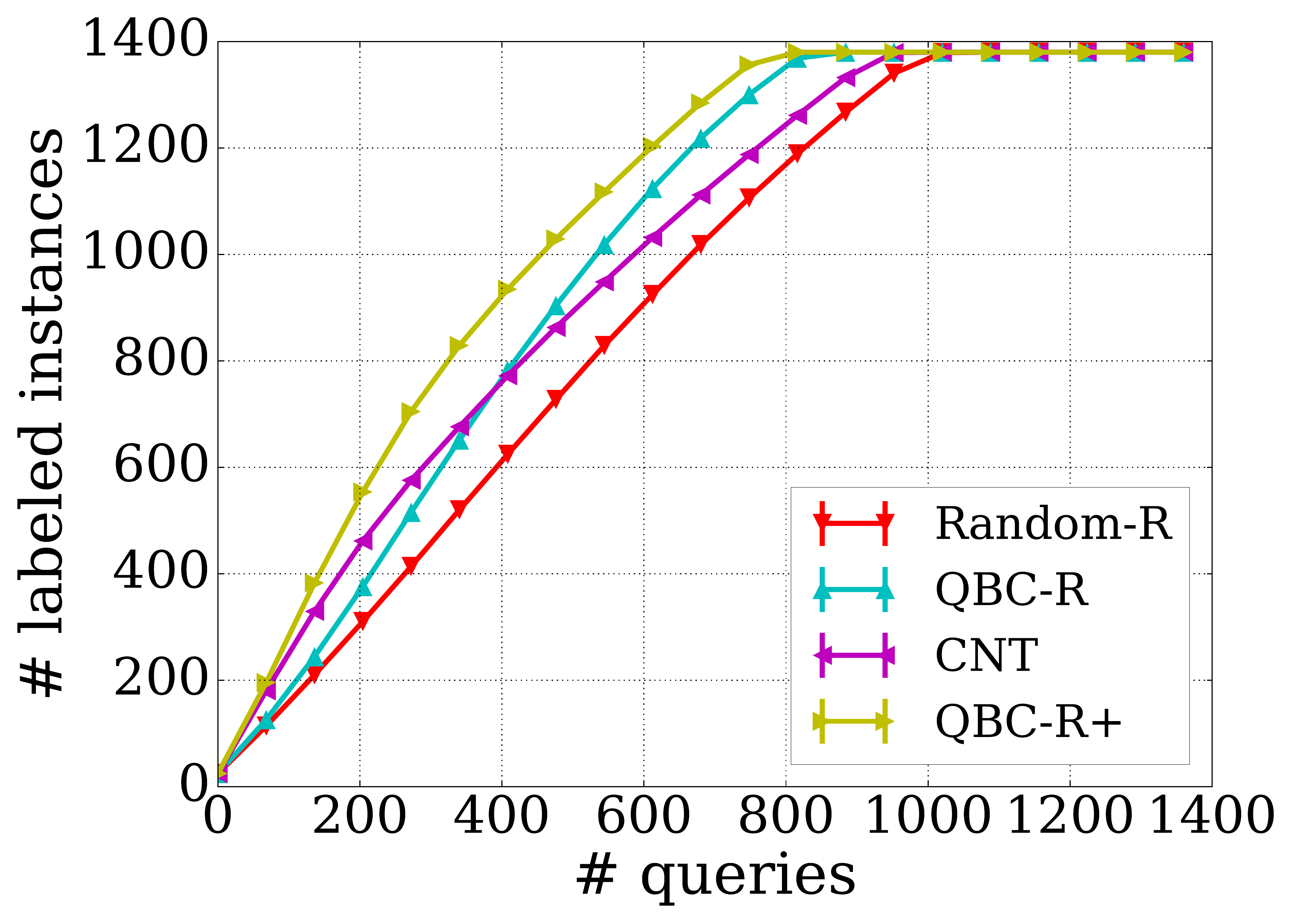}
}
\vspace{-0.3cm}
\caption{Comparison of relations between the number of queries and the number of labeled instances when using different query strategies.}
\vspace{-0.3cm}
\label{fig:size}
\end{figure*}

Figure~\ref{fig:results} shows the AUC results of different query strategies.
For each case, we present the average values and 95\% C.I. of repeated 300 trials with different train/test splits. In addition, Figure~\ref{fig:size} compares the relations between the number of queries and the number of labeled instances across different query strategies. Note that in the relational active learning setting querying a single unlabeled instance will result in one or more labeled instances. 
According to Figure~\ref{fig:results} and Figure~\ref{fig:size}, we have the following observations:

% 1. Relation helps
First, by comparing query strategies under the settings (ii) and (iii) with setting (i), we observe that incorporating relational reasoning into active learning substantially improves the AUC performance of each query strategy. In addition, we find the query order, which is supposed to be different for each strategy, does not affect $\mathcal D_l$ at the end when $\mathcal D\subseteq \mathcal D_l$. Thus, it partly verifies Corollary~\ref{cor:consistent}.
% 2. -R+ is better than -R
Second, our proposed LC-R+ and QBC-R+ significantly outperform other compared query strategies. Specifically, when comparing them with LC-R and QBC-R, we see that incorporating relational reasoning into directing the queries helps to train a better classifier. Figure~\ref{fig:size} shows that LC-R+ and QBC-R+ lead to more labeled instances when using the same amount of queries than that of LC-R and QBC-R. This partly contributes to the performance gain.
% 3. -R+ is more effective at increasing labeled set
Third, LC-R+ and QBC-R+ are more effective at both collecting a larger labeled set and training better classifiers than the CNT baseline. In addition, by comparing CNT with LC-R, QBC-R, and Random-R, we observe that a larger size of the labeled set does not always lead to a better performance. Such observations demonstrate the necessity of combining deterministic relational reasoning and probabilistic machine learning in designing query strategies.

\begin{figure*}[ht!]
\centering
\subfigure[Data Mining]{
\includegraphics[width=\fw\textwidth]{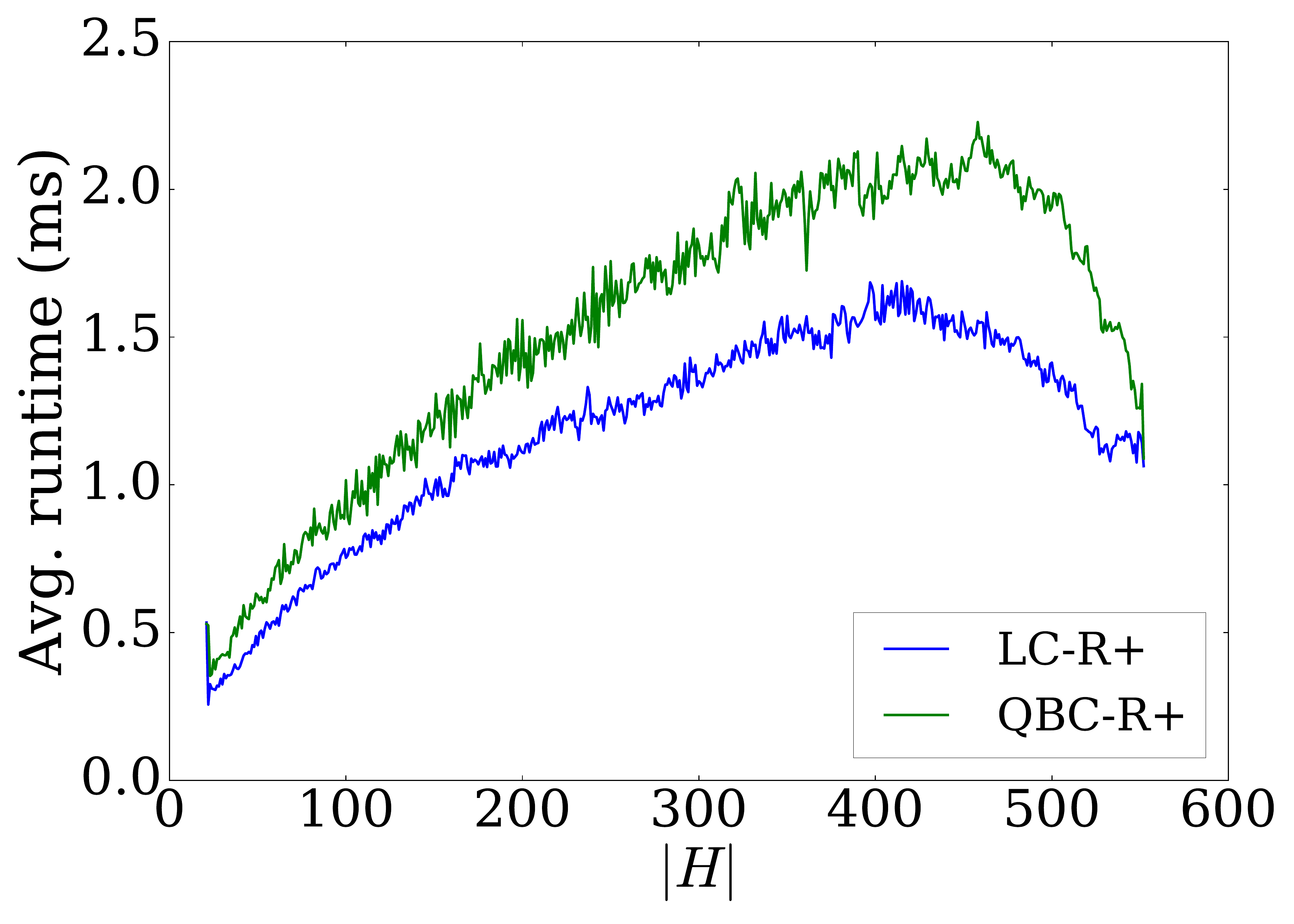}
}
\subfigure[Geometry]{
\includegraphics[width=\fw\textwidth]{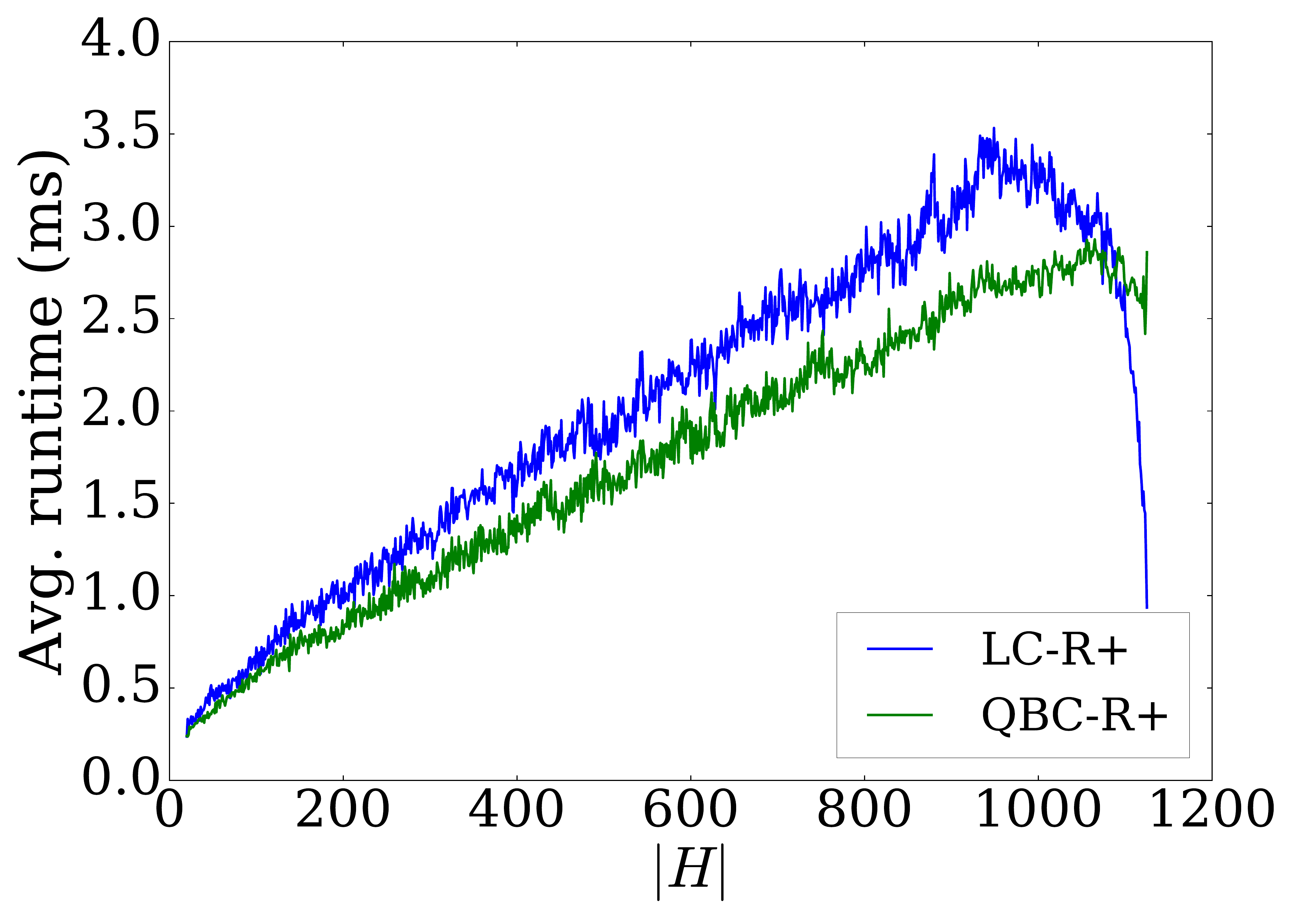}
}
\subfigure[Physics]{
\includegraphics[width=\fw\textwidth]{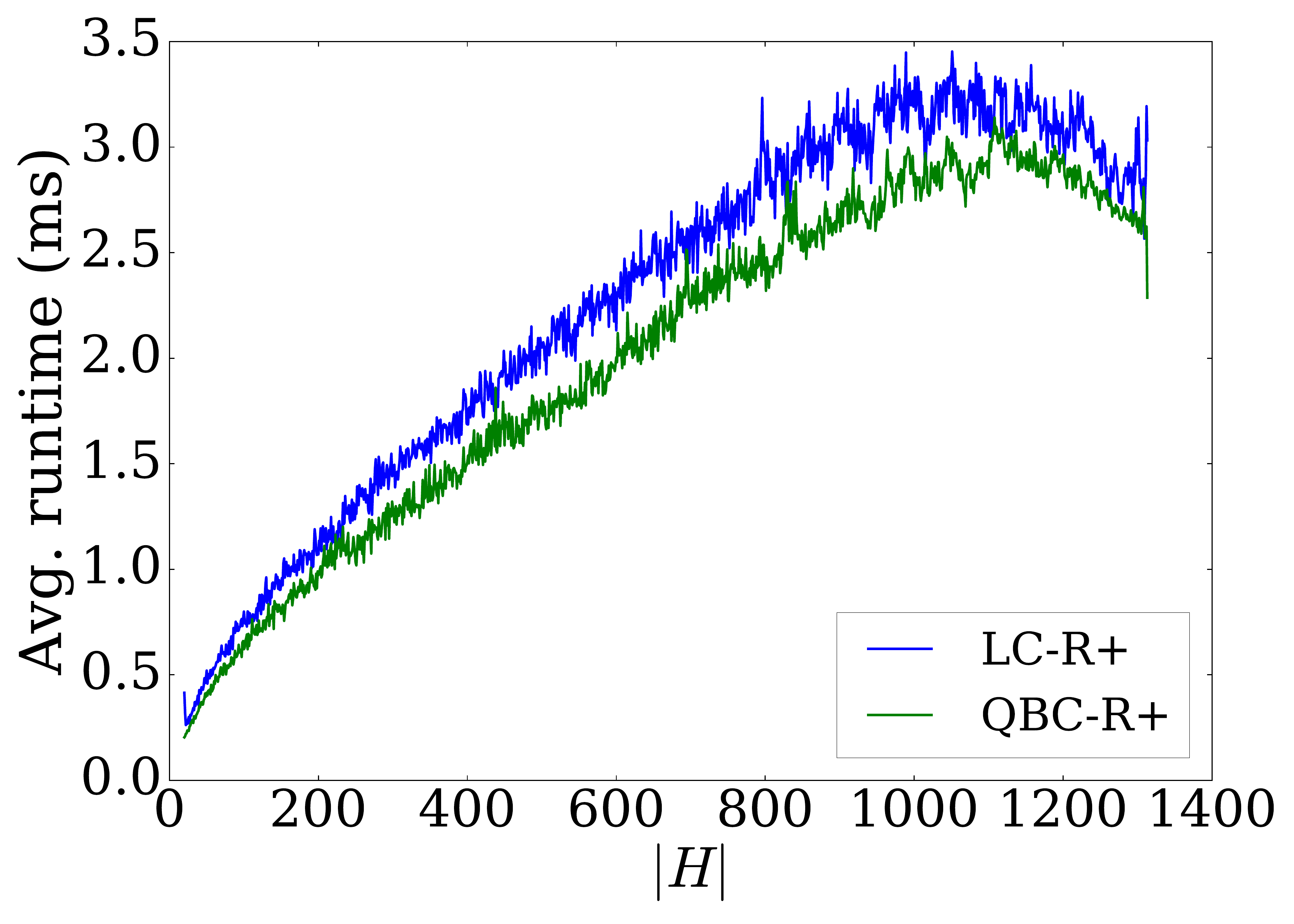}
}
\subfigure[Precalculus]{
\includegraphics[width=\fw\textwidth]{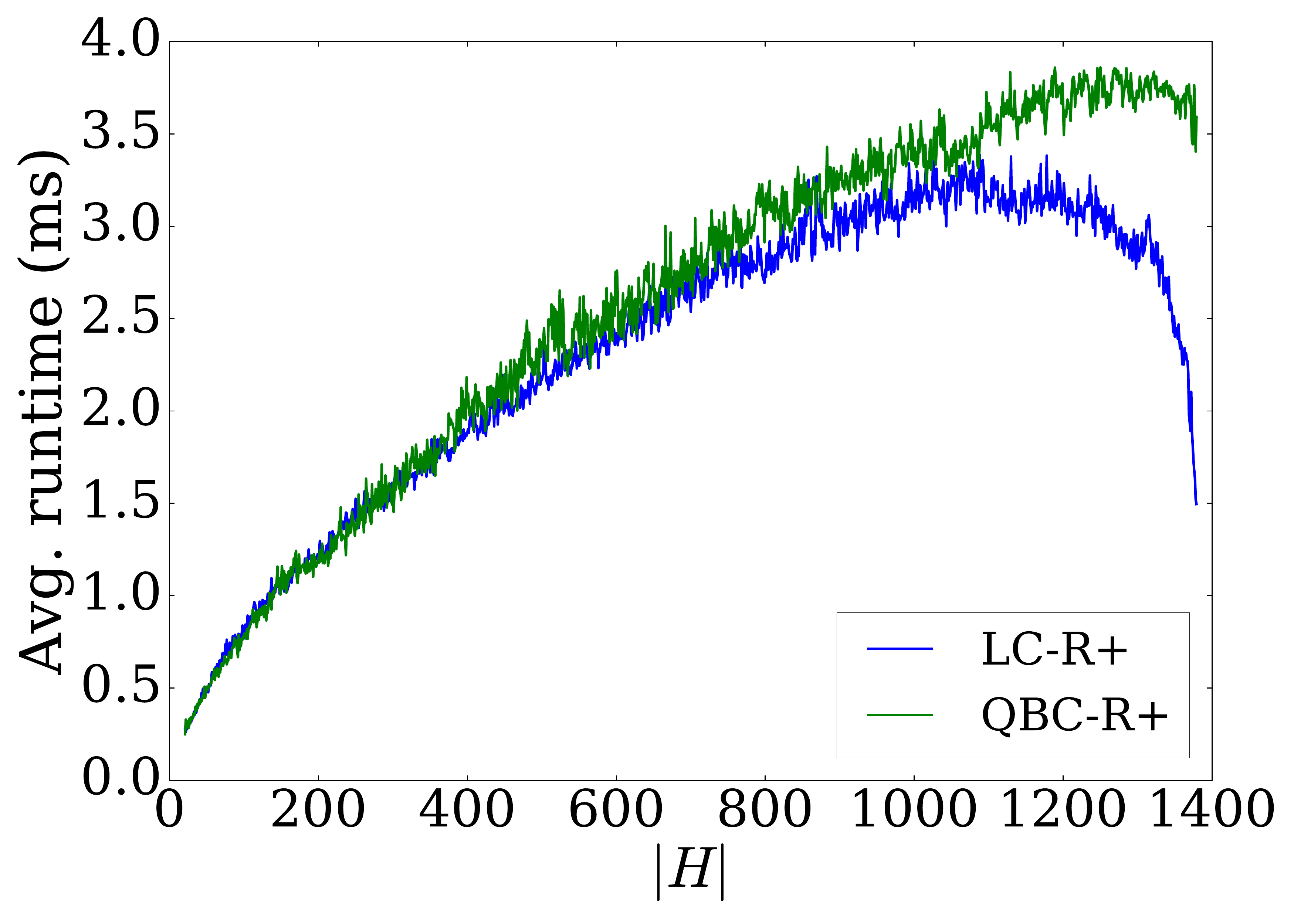}
}
\vspace{-0.3cm}
\caption{The average runtime for calculating the new closure using Theorem~\ref{thm:cxy} v.s. the size of current labeled closure $|H|$. }
\vspace{-0.6cm}
\label{fig:runtime}
\end{figure*}

\subsubsection{Efficiency Study}
\label{sec:efficiency}
The proposed reasoning module is designed to be plugged into any algorithm that needs reasoning of strict orders. 
Thus besides verifying effectiveness, it is also important to investigate its efficiency. 
We conduct empirical studies on the runtime of the reasoning module.

Figure~\ref{fig:runtime} shows the relation between the average runtime for calculating the new closure using Theorem~\ref{thm:cxy} and the size of the current labeled closure $|H|$. Results for both LBC-R+ and QBC-R+ are presented. We can see that as $|H|$ keeps increasing during the pool-based active learning process, the average runtime of calculating $C_{(x,y)}$ increases almost linearly and even decreases a little at the end. 
Although the worst case time complexity for calculating Theorem~\ref{thm:cxy} is $O(|H|^3)$ (for $O_{(a,b)}$) and $O(|H|^2)$ (for others), the runtime required is directly related to the number of ascendants and descendants of elements in $V$, which is usually different for the four strict order datasets used.
If the few ascendants and descendants effectively control the size of calculations as we have observed, the runtime will be short regardless of a large $|H|$.
This might explain why the growth of calculating $C_{(x,y)}$ is near linear.
We also empirically evaluate the effects of using Prop.~\ref{prop:efficiency} on the efficiency and include the results in the supplemental material.

\vspace{-0.2cm}
\section{Conclusion}
\label{sec:conclusion}
\vspace{-0.2cm}
We propose an active learning framework tailored to relational data in the form of strict partial orders.
An efficient reasoning module is proposed to extend two commonly used query strategies -- uncertainty sampling and query by committee.
Experiments on concept prerequisite learning show that incorporating relational reasoning in both selecting valuable examples to label and expanding the training set significantly improves standard active learning approaches.
Future work could be to explore the following: (i) apply the reasoning module to extend other query strategies; 
% (ii) test the proposed algorithm on other types of strict partial orders in which active learning is important; 
(ii) active learning of strict partial orders from a noisy oracle.

\bibliography{acl2018}
\bibliographystyle{acl_natbib}

\clearpage

\appendix
\section{Supplemental Material}

\subsection{Proof of Proposition~\ref{prop:uniq}}
\begin{proof}
For any two supersets $H_1\neq H_2$ of $H$ whose oracles $W_{H_1},W_{H_2}$ are complete,
$W_{H_1\cap H_2}$, on a smaller set $H_1\cap H_2$, is also complete (by definition).
\end{proof}

\subsection{Proof of Proposition~\ref{prop:first}}
\begin{proof}
For any $(a,b),(b,c)\in H\cap G$, because $W_H$ is complete, $(a,c)\in H\cap G$ (by Definition~\ref{def:closure} (i)). For any $(a,b)\in H\cap G$, $(b,c)\in H\cap G^c\subseteq (H\cap G)^c$ (by Definition~\ref{def:closure} (iv)). Therefore, $H\cap G$ is also a strict order of $V$ (by Definition~\ref{def:sorder}).  
\end{proof}

\subsection{Proof of Theorem~\ref{thm:cxy}}\label{sec:proof}

\subsubsection{Well-definiteness}
It is trivial that if $W_H$ is complete, $G\cap (H\cup N_{(x,y)})$
is also a strict order. Therefore, $N''_{(c,d)}$ is well defined, so is $O_{(x,y)}$.

\subsubsection{Necessity}
One can easily verify that if $(x,y)\in G\cap H^c$, both $N_{(x,y)}\subseteq \overline{H'}$ (Definition~\ref{def:complete} (i)), $R_{(x,y)}\subseteq \overline{H'}$ (Definition~\ref{def:complete} (iv)), and
$S_{(x,y)}\cup T_{(x,y)}\cup O_{(x,y)}\subseteq \overline{H'}$ (Definition~\ref{def:complete} (ii),(iii)) from
the definition of closure, and likewise if $(x,y)\not\in G$, $N'_{(x,y)}\subseteq \overline{H'}$
(Definition~\ref{def:complete} (ii), (iii)).
In another word, $C_{(x,y)}(H)\subseteq \overline{H'}$.
Also, see Fig.~\ref{fig:reasoning} for the explanation of each necessary condition mentioned.

\subsubsection{Sufficiency}
One can see $A_{a'}^G\subseteq A_a^G$ if $a'\in A_a^G$ and $D_{a'}^G\subseteq D_a^G$ if $a'\in D_a^G$.
That is, an ancestor of ancestor is also an ancestor (briefly, AAA), and a descendant of descendant is also a descendant (briefly, DDD).

Now we proceed to prove $C_{(x,y)}$ is complete using contradiction which finalizes the proof of our result $C_{(x,y)}=\overline{H'}$.
If $C_{(x,y)}$ is not complete, by definition, one of the four conditions in Definition~\ref{def:complete} must fail.

If Definition~\ref{def:complete} (i) fails, there must exist $a,b,c$ such that $(a,b)\in C_{(x,y)}\cap G$, $(b,c)\in C_{(x,y)}\cap G$,
while $(a,c)\not\in C_{(x,y)}$.
In this case, if both $(a,b)$ and $(b,c)$ are from $H$, because $H$ is complete, $(a,c)\in H\cap G$ contradicts the assumption. Hence
at least one of $(a,b)$ and $(b,c)$ is not included in $H$. By the definition of $C_{(x,y)}$, if one pair belongs to $G\cap H^c$,
it must come from $N_{(x,y)}$. Therefore, it implies $(x,y)\in G\cap H^c$.

\noindent \underline{Cases 1: If $(a,b)\in N_{(x,y)}$ and $(b,c)\in N_{(x,y)}$}, $(y,b)\in G$ and $(b,x)\in G$. That however implies $(y,x)\in G$, contradicting $G$'s definition as a strict order (See Definition~\ref{def:sorder} (ii)).

\noindent \underline{Cases 2: If {$(a,b)\in N_{(x,y)}$ and $(b,c)\in H$}}, $a\in A_x^{G\cap H}$ and $c\in D_y^{G\cap H}$ (by DDD). It implies $(a,c)\in N_{(x,y)}\subseteq C_{(x,y)}$.

\noindent \underline{Cases 3: If {$(a,b)\in H$ and $(b,c)\in N_{(x,y)}$}}, $a\in A_x^{G\cap H}$ (by AAA) and $c\in D_y^{G\cap H}$. It implies $(a,c)\in N_{(x,y)}\subseteq C_{(x,y)}$.

In summary, Definition~\ref{def:complete} (i) holds for $C_{(x,y)}$.

If Definition~\ref{def:complete} (ii) fails, there must exist $a,b,c$ such that $(a,b)\in C_{(x,y)}\cap G = G\cap (H\cup N_{(x,y)})$, $(a,c)\in C_{(x,y)}\cap G^c$,
while $(b,c)\not\in C_{(x,y)}$. In this case, if both $(a,b)$ and $(a,c)$ are from $H$, because $H$ is complete, $(b,c)\in H\cap G^c$ contradicts the assumption. Hence {\em at least one of $(a,b)$ and $(a,c)$ is not included in $H$}.
We divide the statement into the following cases to discuss:

\noindent \underline{Cases 1: If $(x,y)\in G$, $(a,b)\in N_{(x,y)}$, and } \underline{$(a,c)\in H\cap G^c$},
$(a, x) \in {G\cap H}$ and $(y,b) \in {G\cap H}$. Because $(a, x) \in {H\cap G}$,
and $(a,c)\in H\cap G^c$, $(x,c)\in H\cap G^c$. Because $\{(x,y),(y,b)\}\subseteq G\cap (H\cup N_{(x,y)})$,
$(x,b)\in G\cap (H\cup N_{(x,y)})$. Therefore, $(b,c)\in N''_{(x,c)}\subseteq C_{(x,y)}$.
% , which contradicts the assumption.

\noindent \underline{Cases 2: If {$(x,y)\in G$ and $(a,c)\in R_{(x,y)}$}},
$(c,a)\in N_{(x,y)}$, thus $(c,b)\in N_{(x,y)}$. It implies $(b,c)\in R_{(x,y)}\subseteq C_{(x,y)}$.
% which again contradicts the assumption.

\noindent \underline{Cases 3: If {$(x,y)\in G$ and $(a,c)\in S_{(x,y)}$}}, $(y,a)\in G\cap H$ and $\exists (d,c)\in G^c\cap H$ such that $(d,x)\in G\cap H$. Thus, $(x,c)\in G^c\cap H\Rightarrow (y,c)\in S_{(x,y)}\Rightarrow (b,c)\in N''_{(y,c)}\subseteq C_{(x,y)}$.
% which again contradicts the assumption.

\noindent \underline{Cases 4: If {$(x,y)\in G$ and $(a,c)\in T_{(x,y)}$}}, $(c,x)\in G\cap H$ and $\exists (a,d)\in G^c\cap H$ such that $(y,d)\in G\cap H$. Thus, $(a,y)\in G^c\cap H\Rightarrow (a,x)\in T_{(x,y)}\Rightarrow (a,b)\not\in N_{(x,y)}$. Because $(a,b)\in G\cap (H\cup N_{(x,y)})$, one has
$(a,b)\in G\cap H\Rightarrow (b,x)\in T_{(x,y)}\Rightarrow (b,c)\in N''_{(b,x)}\subseteq C_{(x,y)}$.
% This again contradicts the assumption.

\noindent \underline{Cases 5: If {$(x,y)\in G$ and $(a,c)\in O_{(x,y)}$}}, there exists $(d,e)\in S_{(a,b)}\cup T_{(a,b)}$ such that $(a,c)\in N''_{(d,e)}$.
Therefore, $\{(a,b),(d,a),(c,e)\}\subseteq G\cap (H\cup N_{(x,y)})$.
Hence, $(b,c)\in N''_{(d,e)}\subseteq C_{(x,y)}$.

\noindent \underline{Cases 6: If {$(x,y)\in G^c$}}, we have $(a,b)\in G\cap H$ and $(a,c)\in N'_{(x,y)}$.
Thus $(x,b)\in G\cap H \Rightarrow (b,c) \in N'_{(x,y)}\subseteq C_{(x,y)}$.

In summary, all six cases above contradict the assumption $(b,c)\not\in C_{(x,y)}$. Thus Definition~\ref{def:complete} (ii) holds for $C_{(x,y)}$.
Given we have verified the condition of Definition 3 (ii), one can also prove that Definition 3 (iii) holds in a similar way, because their statements as well as definitions of $S_{(x,y)}$ and $T_{(x,y)}$ are symmetric.

One can easily see that Definition 3 (iv) holds for $C_{(x,y)}$, because $G\cap (H\cup N_{(x,y)})$ is also a strict order and $R_{(x,y)}\subseteq C_{(x,y)}$. \QED

\subsection{Proof of Proposition~\ref{prop:complexity}}
\begin{proof}
If $(c,d) \in S_{(a,b)}$, $c\in D_{b}^{G\cap H}\cup \{b\}$.
then $c\not\in A_{a}^{G\cap H}\cup \{a\}$, thus
$D_{c}^{G\cap (H\cup N_{(a,b)})} = D_{c}^{G\cap H}$ (by
Definition of $N_{(a,b)}$). It holds that
$D_{c}^{G\cap (H\cup N_{(a,b)})}\cup \{c\}\subseteq D_{b}^{G\cap H}\cup \{b\}$.
Therefore, one has $N''_{(c,d)}\subseteq N''_{(b,d)}\subseteq O_{(a,b)}$ if $(c,d) \in S_{(a,b)}$.
Likewise, $N''_{(c,d)}\subseteq N''_{(c,a)}\subseteq O_{(a,b)}$ if $(c,d)\in T_{(a,b)}$. One has
\[O_{(a,b)} = \left[\bigcup_{(c,d)\in S_{(a,b)}} N''_{(b,d)} \right]\cup \left[
\bigcup_{(c,d)\in T_{(a,b)}} N''_{(c,a)}\right],\] whose time complexity is $O(|H|^3)$.
\end{proof}

\subsection{Proof of Proposition~\ref{prop:efficiency}}
\begin{proof}
Let $(c',d')\in N''_{(c,d)}$.
If $d'\neq d$, $(d',d)\in G\cap(H\cup N_{(a,b)})$. Then, $A_{d'}^{G\cap(H\cup N_{(a,b)})}\subseteq A_{d}^{G\cap(H\cup N_{(a,b)})}$ (AAA). Thus, 
\[A_{d'}^{G\cap(H\cup N_{(a,b)})}\cup\{d'\}
\subseteq A_{d}^{G\cap(H\cup N_{(a,b)})}\cup\{d\}.\]
Likewise, 
\[D_{c'}^{G\cap(H\cup N_{(a,b)})}\cup\{c'\}\in D_{c}^{G\cap(H\cup N_{(a,b)})}\cup\{c\}.\]
By the definition of $N''_{(c',d')}$ and $N''_{(c,d)}$, one has $N''_{(c',d')}\subseteq N''_{(c,d)}$.
\end{proof}

\subsection{Proof of Theorem~\ref{thm:bound}}
We first introduce the notion of \emph{transitive reduction} before we proceed:
\begin{definition}[Transitive Reduction~\cite{aho1972transitive}]
Let G be a directed acyclic graph. We say \underline{G} is a transitive reduction of G if:
\begin{enumerate}
\item[(i)] There is a directed path from vertex u to vertex v in \underline{G} iff there is a directed path from u to v in G, and
\item[(ii)] There is no graph with fewer arcs than \underline{G} satisfying (i).
\end{enumerate}
\end{definition}

For directed acyclic graph $G$, \citet{aho1972transitive} have shown that the transitive reduction is unique and is a subgraph of $G$. Let $G$ be a simple directed acyclic graph (DAG). In compliance with Def.~\ref{def:closure}, we use $\overline{G}$ to denote the transitive closure of $G$. Define $S(G)$ as the set of graphs such that every graph in $S(G)$ has the same transitive closure as $G$, i.e.,
\begin{equation*}
S(G)\coloneqq\{G^\prime\mid\overline{G^\prime}=\overline{G}\}
\end{equation*}

\citet{aho1972transitive} have shown that $S(G)$ is closed under intersection and union. Further more, for DAG $G$, the following relationship holds:
\begin{equation*}
\underline{G}=\bigcap_{G^\prime\in S(G)}G^\prime \subseteq G \subseteq \bigcup_{G^\prime \in S(G)} G^\prime = \overline{G}
\end{equation*}

Next, we give the proof of Theorem~\ref{thm:bound} below.
\begin{proof}
With the fact that negative labels from the query oracle cannot help to induce positive labels in the graph, we can bound the number of queries, $m$, needed to learn a classifier:
$$m\geq |\underline{G}|$$
The proof is by simple contradiction based on the definition of the transitive reduction of $G$ and the fact that $A$ is a consistent learner. On the other hand, there is a learning algorithm $A$ that simply remembers all the queries with positive labels and predict all the other inputs as negative. For this algorithm $A$, it suffices for $A$ to make $|\overline{G}|$ queries.
\end{proof}

\subsection{Experiment Environment}
All experiments are conducted on an Ubuntu 14.04 server with 256GB RAM and 32 Intel Xeon E5-2630 v3 @ 2.40GHz processors.
Active learning query strategies are implemented in Python2.7.
Code and data will be publicly available.

\subsection{Effectiveness of Proposition~\ref{prop:efficiency}}
\begin{table}[tp]
\centering
\scalebox{0.67}{
\begin{tabular}{lcccc}
\toprule
\textbf{Domain} & \multicolumn{2}{c}{\textbf{LC-R+}} & \multicolumn{2}{c}{\textbf{QBC-R+}}\\
 & \textbf{w/o pruning} & \textbf{w/ pruning} & \textbf{w/o pruning} & \textbf{w/ pruning}\\
\midrule
Data mining & 5.5   & 3.3 (-40\%)   & 5.5  & 2.6 (-53\%)\\
Geometry    & 85.5  & 39.2 (-54\%)  & 69.6 & 31.4 (-55\%)\\
Physics     & 111.0 & 58.8 (-47\%)  & 117.1& 60.2 (-49\%)\\
Precalculus & 134.3  & 59.9 (-55\%) & 167.4& 74.8 (-55\%)\\
\bottomrule
\end{tabular}
}
\caption{The effect of Proposition~\ref{prop:efficiency} on the total runtime (s) of $O_{(a,b)}$ calculation in Theorem~\ref{thm:cxy} for a full round of active learning (until $\mathcal D_u=\emptyset$).}
\label{tab:prune}
% \vspace{-0.5cm}
\end{table}

We also empirically evaluate the effects of using Proposition~\ref{prop:efficiency} on the efficiency. Specifically, we measure the total runtime of $O_{(a,b)}$ calculation in Theorem~\ref{thm:cxy} for a full round of active learning (until $\mathcal D_u=\emptyset$) with and without applying the pruning rule induced by Proposition~\ref{prop:efficiency}. The results are shown in Table~\ref{tab:prune} where the numbers are the average runtime over all 300 different rounds of active learning for each dataset. We can see that the pruning can lead to a speedup of 40-55\% in the LC-R+ experiments and a speedup of 49-55\% in the QBC-R+ experiments, which shows that Proposition~\ref{prop:efficiency} is helpful for higher efficiency.

\end{document}